\documentclass{article}

    \usepackage[final]{neurips_2025}

\usepackage[utf8]{inputenc} 
\usepackage[T1]{fontenc}    
\usepackage{hyperref}       
\usepackage{url}            
\usepackage{booktabs}       
\usepackage{amsfonts}       
\usepackage{nicefrac}       
\usepackage{microtype}      
\usepackage{xcolor}         

\usepackage{xstring}
\usepackage{multirow}
\usepackage{makecell} 
\usepackage{tabularx} 

\usepackage{amsfonts, amsmath, amssymb, bm, bbm, amsthm}
\usepackage{subcaption}
\usepackage{graphicx}

\newcommand{\R}{\mathbb{R}}
\newcommand{\E}{\mathbb{E}}

\newcommand{\appref}[1]{\ref{#1}}

\DeclareMathAlphabet{\mathmybb}{U}{bbold}{m}{n} 

\usepackage{thmtools}
\usepackage{thm-restate}
\usepackage{hyperref}
\usepackage{cleveref}
\declaretheorem[name=Theorem]{theorem}
\declaretheorem[name=Lemma, sibling=theorem]{lemma}
\declaretheorem[name=Proposition]{proposition}

\newcounter{relctr} 
\everydisplay\expandafter{\the\everydisplay\setcounter{relctr}{0}} 

\newcommand\labelrel[2]{%
  \begingroup
    \refstepcounter{relctr}%
    \stackrel{\textnormal{(\roman{relctr})}}{\mathstrut{#1}}%
    \originallabel{#2}%
  \endgroup
}
\AtBeginDocument{\let\originallabel\label} 
\newcommand{\propDecomposed}[1]{%
    \begin{proposition}
    Assume $h:\{0, 1\}^n \rightarrow \R$ can be decomposed as follows: $h(x) = \sum\limits_{i=1}^{p} h_i(x_{S_i}), S_i \subseteq [n]$.
    That is, each function $h_i: \{0,1\}^{|S_i|} \rightarrow \R$ depends on at most $|S_i|$ variables.  Then,  $h$ is $k=O( \sum\limits_{i=1}^p 2^{|S_i|})$-sparse and of degree $d=\max(|S_1|, \dots, |S_p|)$.
    \end{proposition}
}

\newcommand{\propDegree}[1]{%
    \begin{proposition}
    Let, $h:\{0,1\}^n \rightarrow \R$ be pseudo-boolean function and let $d \in \mathbb{N}$ be some constant (w.r.t. $n$). If $h$ is of degree $d$, then, it is $k=O(n^d)$-sparse.
    \end{proposition}
}

\title{SHAP values via sparse Fourier representation}

\author{Ali Gorji \thanks{Equal contribution.} \\
ETH Zürich, Switzerland \\
\texttt{ali.gorji@alumni.ethz.ch} \\
\And
Andisheh Amrollahi $^*$\\
ETH Zürich, Switzerland \\
\texttt{amrollaa@alumni.ethz.ch} \\
\And
Andreas Krause \\
ETH Zürich, Switzerland \\
\texttt{krausea@ethz.ch}
}

\begin{document}

\maketitle

\begin{abstract}
\vspace*{-2mm}

SHAP (SHapley Additive exPlanations) values are a widely used method for local feature attribution in interpretable and explainable AI. We propose an efficient two-stage algorithm for computing SHAP values in both black-box setting and tree-based models. We assume the black-box predictor or tree model accepts binary (zero-one) features.
Motivated by spectral bias in real-world predictors, we first approximate the predictor using compact Fourier representations, exactly for trees and approximately for black-box models. In the second stage, we introduce a closed-form formula for {\em exactly} computing SHAP values using the Fourier representation, that ``linearizes'' the computation into a simple summation and is amenable to parallelization. As the Fourier approximation is computed only once, our method enables amortized SHAP value computation, achieving significant speedups over existing methods and a tunable trade-off between efficiency and precision.
\end{abstract}
\vspace*{-2mm}
\section{Introduction}
\vspace*{-2mm}
\label{sec:intro}
Interpretability of machine learning models is paramount, especially in high-stakes applications in areas such as medicine, 
fraud detection, 
or credit scoring.
This is crucial to the extent that in Europe, the General Data Protection Regulation (GDPR) mandates the \emph{legal} right to an explanation of algorithmic decisions \citep{voigt2017eu}. Say we are given a predictor/model $h:\mathcal{X}^n \rightarrow \mathbb{R}$ which maps an input (data) instance $x^* \in \mathcal{X}^n$ to a prediction $h(x^*)$. \emph{Instance-wise} a.k.a. \emph{local} feature attribution methods assign ``importances'' to each of the features $x_i^*  \in \mathcal{X}$ of the instance $x^*$ which quantify how influential that feature was in the model predicting the value $h(x^*)$.

A widely used method for deriving attributions (importances) is the notion of \emph{SHapley Additive exPlanations}, commonly referred to simply as SHAP values. Originally, the notion of Shapley values was introduced in the seminal work of \citet{P-295} in the context of cooperative game theory. 
The Shapley value is a mathematically well-founded and ``fair'' way of distributing a reward among all the members of a group playing a cooperative game and it is computed based on the rewards that would be received for all possible coalitions. The Shapley value is the unique way of distributing the reward that satisfies several reasonable mathematical properties that capture a notion of fairness \citep{P-295}. In the context of machine learning and statistics, the players become features, the reward is the prediction of the predictor $h$ and the SHAP value is the ``contribution'' or ``influence'' of that feature on the prediction. Shapley values are widely used due to their mathematical soundness and desirable properties (\cite{gromping2007estimators,  vstrumbelj2009explaining, owen2014sobol, datta2016algorithmic, owen2017shapley,   lundberg2017unified, lundberg2020local} \cite{aas2021explaining}).

\looseness -1
Despite their prevalence, computing SHAP values is challenging, as it involves an exponential summation, i.e., a summation over exponentially many terms , see Equation~\ref{eqn:shap}. This is because the formula accounts for a feature's importance in the context of all possible ``coalitions'' of other features, and therefore the formula covers all possible subsets of other features. 
Therefore, approximating them and speeding up the computation has received attention in a variety of settings. SHAP value computation can easily dominate the computation time of industry-level machine learning solutions on datasets with millions or more entries \citep{yang2021fast}. \citet{yang2021fast} point out that industrial applications sometimes require hundreds of millions of samples to be explained. Examples include feed ranking, ads targeting, and subscription propensity models. In these modeling pipelines, spending tens of hours in model interpretation becomes a significant bottleneck \citep{yang2021fast} and one usually needs to resort to multiple cores and parallel computing.

Significant work has gone into speeding up the computation of SHAP values for a variety of settings. In the (ensemble of) trees setting, full access to the tree structure is assumed. \citet{yang2021fast, bifet2022linear} provide theoretical and practical computational speedups to the classic \textsc{TreeShap} \citep{lundberg2020local}. Similar to these results, in this work, we provide significant speedups for the tree setting over previous methods. 

As opposed to the tree setting, which is a ``white box'' setting, in the {\em model-agnostic} a.k.a {\em black-box} setting, we only have \emph{query access} to the model. Here, our only means of access to the predictor is that  we can pick an arbitrary $x \in \mathcal{X}^n$ and query the predictor for its value $h(x)$. The usual approach here is to approximate the exponential sum of the SHAP value computations using stochastic sampling \citep{covert2020improving, mitchell2022sampling, lundberg2017unified}. In this setting, \citet{covert2020improving, mitchell2022sampling} provide sampling methods that require fewer queries to the black-box compared to vanilla \textsc{KernelShap} \citep{lundberg2017unified} for equal approximation accuracy. \textsc{FastShap} \citet{jethani2021fastshap}, which introduces a method for estimating Shapley values in a single forward pass using an end-to-end learned explainer neural network model. Our algorithm \textsc{FourierShap} falls into the query-access black-box setting as well. However, we take a different approach. We are guided by the key insight that many models used in practice have a ``spectral bias''. \citet{yang2019fine, valle2018deep} provably and experimentally show that fully connected neural networks with binary (zero-one) inputs learn low-degree -- and therefore sparse -- functions in a basis called the \emph{Walsh-Hadamard a.k.a Fourier} basis. It is well known that the Walsh-Hadamard transform (WHT) of an ensemble of $T$ trees of depth $d$ is also of degree at most $d$ and moreover, $k=O(T4^d)$-sparse \citep{kushilevitz1993learning, mansour1994learning}.

\textbf{Our contributions:}  
Guided by the aforementioned insights, we provide an algorithm to compute SHAP values using the Fourier representation of the tree or black-box model. We first approximate the black-box function by taking its sparse Fourier transform. We theoretically justify, and show through extensive experiments, that for many real-world models such as fully connected neural networks and (ensembles of) trees this representation is accurate. Subsequently, we prove that SHAP values for a single Fourier basis function admit a closed-form expression not involving an exponential summation. Therefore, using the Fourier representation we overcome the exponential sum and can utilize compute power effectively to compute SHAP values. Furthermore, the closed-form expression we derive effectively ``linearizes'' the computation into a simple summation and is amenable to parallelization on multiple cores or a GPU. The Fourier approximation step is only done {\em once}, therefore  \textsc{FourierShap} amortizes the cost of computing explanations for many inputs. Subsequently, SHAP computations using the Fourier approximation are orders of magnitude faster compared to other black-box SHAP approximation methods such as \textsc{KernelShap} and other variations of it which all involve a computationally expensive optimization. We show  speedups over other methods such as \textsc{DeepLift} \citep{shrikumar2017learning} and \textsc{FastShap} \citep{jethani2021fastshap} as well. In addition to the speedup, our algorithm enables a reliable continuous trade-off between computation and accuracy, controlled in a fine manner by the sparsity of the Fourier approximation. 

\section{Background}
\vspace*{-2mm}
\label{sec:background}
This section reviews the notion of SHAP values, and sparse and low-degree Fourier transforms. 
\subsection{Shapley values} 
\vspace*{-1mm}
In game theory, a \emph{cooperative game} is a function $v: 2^{[n]} \rightarrow \R$ that maps a subset (coalition) $S \subseteq [n]$ of a group of players $[n]=\{1, \dots, n\}$ to their total reward (when they cooperate). If all the players cooperate they win a total reward of value $v([n])$. The main question is how they would distribute this reward among themselves. \citet{P-295} resolved this problem by a deriving a \emph{unique} value based on ``fairness axioms'' proposed in his seminal work \citep{P-295}. The \emph{Shapley value} of player $i \in [n]$ is:
\begin{equation}
\label{eqn:shap}
\phi_i(v) = \frac{1}{n} \sum\limits_{S \subseteq [n] \setminus \{i\}} \frac{v(S \cup\{i\}) - v(S)}{  \binom{n-1}{|S|}}
\end{equation}

Intuitively, one can view the term $v(S \cup\{i\}) - v(S)$ as the marginal contribution of player $i$ when they are added to the coalition $S$. This marginal value is weighted according to the number of permutations the leading $|S|$ players and trailing $n-|S|-1$ players can form. 

\looseness -1 In the machine learning context, we have a predictor $h: \mathcal{X}^n \rightarrow \R$ mapping an $n$-dimensional feature vector to a value. In this context, the players become features $x_i \in \mathcal{X}$ and the reward is the prediction of the predictor $h$ and the Shapley value is the ``contribution'' or ``influence'' of the $i$'th feature on the prediction. We define $v(S)$ accordingly to capture this notion \citep{lundberg2017unified}:
\[
v(S) = \E_{\bm{x_{[n] \setminus S}} \sim p(\bm{x_{[n] \setminus S}}))}[h(x^*_S, \bm{x_{[n] \setminus S}})],
\]
where $x^* \in \mathcal{X}^n$ is the instance we are explaining. 
This definition implicitly captures the way we handle the missing features (feature not present in the coalition): we integrate the missing features with respect to the marginal distribution $p(\bm{x_{[n] \setminus S}}))$. In practice, the marginalization is performed with an empirical distribution by taking a subset of the training data as \emph{background dataset}.  

The choice of which distribution to average the missing features from has been discussed thoroughly in the relevant literature. As mentioned before, in this work, we focus on the SHAP values as defined in \textsc{KernelShap} introduced by \citet{lundberg2017unified,lundberg2020local} also known as  ``Interventional'' \citep{pmlr-v108-janzing20a,  van2021tractability} or ``Baseline''  \citep{sundararajan2020many} SHAP, where the missing features are integrated out from the \emph{marginal distribution} $p(\bm{x_{[n] \setminus S}})$, as opposed to the \emph{conditional distribution} $p(\bm{x_{[n] \setminus S}} | \bm{x_S})$. We refer the reader to Appendix~\appref{subsec:shapley_literature_review} for a comprehensive overview of the literature discussing these two notions and their conceptual differences.

Since we will be using the well-known \textsc{KernelShap} \citep{lundberg2017unified} and its variant \textsc{LinRegShap} \citep{covert2020improving} as a baseline, we briefly review their method here. \citet{lundberg2017unified} propose the ``least squares characterization'' of SHAP values. They prove that SHAP values are the solution to the following minimization problem:
\begin{multline*}
\beta_0^*, \dots, \beta_n^*  \triangleq \arg\min\limits_{\beta_0, \dots, \beta_n}
\sum\limits_{0<|S|<n} \frac{n-1}{{\binom{n}{|S|}}|S|(n-|S|)} \left( \beta_0 + \sum\limits_{i \in S} \beta_i - v(S) \right) 
\\
s.t.: \;\;\; \beta_0 = v(\{\}), \beta_0 + \sum\limits_{i=1}^n \beta_i = v([n])
\end{multline*}
Then $\phi_i(v) = \beta_i^*$. 

\looseness -1 Unfortunately, the above optimization still involves an exponential sum. Therefore, \citet{lundberg2017unified} propose to sample subsets $S$ uniformly at random. \citet{covert2020improving, mitchell2022sampling} provide better ways of sampling and solving the optimization to get approximations with lower variances and biases. Nevertheless, all these methods require solving a least squares minimization subject to constraints for each explained instance $x^*$ and, therefore, are computationally expensive.

Later, we also compare our method to \textsc{FastSHAP}~\citep{jethani2021fastshap}, a model-agnostic algorithm for computing SHAP values.  In \textsc{FastSHAP}, a parametric explainer $\phi$ (e.g., MLP) is trained to directly generate SHAP values given data samples. Since training is only done once, this, similar to us, it amortizes the cost of generating SHAP values across multiple instances. Computing SHAP values only requires a forward pass on the trained model and therefore can be done very quickly.

\vspace*{-1mm}
\subsection{Fourier representations}
\vspace*{-1mm}
Here we review the notions of the Fourier basis and what we mean by sparsity. We will later use sparse Fourier representation of functions to compute SHAP values. In this work, we focus on the setting where the inputs to the black-box function (predictor) are binary, i.e., $\mathcal{X}^n = \{0,1 \}^n$. This means, we assume we have binary features and/or categorical features, through standard one-hot representations Continuous features can be discretized into quantiles, enabling their transformation into categorical features. 
The Fourier representation of the \emph{pseudo-boolean} function $h:\{0,1\}^n \rightarrow \R$ is the unique expansion of $h$ as follows: $h(x) = \frac{1}{\sqrt{2^n}}\sum\limits_{f \in \{0,1\}^n} \widehat{h}(f) (-1)^{\langle f,x \rangle}$.

The inner product of two vectors $f,x \in \{0,1\}^n$ is defined as:
$\langle f, x \rangle \equiv \sum_i^n f_i x_i,   \forall f, x \in \{0, 1\}^n$. 

The unique function $\widehat{h}: \{0,1\}^n \rightarrow \mathbb{R}$ is called the \emph{Fourier transform} of $h$. For any $f \in \{0, 1\}^n$, $\widehat{h}(f)$ is called the Fourier coefficient corresponding to the frequency $f$. The family of functions $\frac{1}{\sqrt{2^n}}\Psi_f(x) = (-1)^{\langle f,x \rangle }, f \in \{0,1\}^n$ are the $2^n$-many Fourier basis functions. These basis functions are orthonormal:  
$
\sum\limits_{x \in \{0, 1\}^n}  \Psi_{f}(x) \Psi_{f'}(x) = 
\begin{cases} 
      0 & f \neq f'    \\
      1  & f=f' 
\end{cases}, \;\;\; f, f'\in \{0, 1\}^n
$. Therefore, they form a basis for the vector space of all pseudo-boolean functions $h:\{0, 1\}^n \rightarrow \mathbb{R}$. 

We define the support of $h$ to be $\text{supp}(h) = \{f \in \{0,1\}^n | \widehat{h}(f) \neq 0\}$. We say that a function $h$ is \emph{$k$-sparse} if at most $k$ of the $2^n$ Fourier coefficients $\widehat{h}(f)$ are non-zero, i.e., $|\text{supp}(h)| \leq k$. The degree of a vector $f \in \{0,1\}^n$ is denoted by $\deg(f)$ and is defined as the number of ones in the vector. For example, if $n=5$ then $f=(1,0,0,1,0)$ is a vector of degree $\deg(f)=2$.  We say a function is \emph{degree $d$} when the frequencies $f \in \{0,1\}^n$ corresponding to non-zero Fourier coefficients are of degree less or equal to $d$ i.e. $\forall f \in \text{supp}(h)$ it holds that $\text{deg}(f) \leq d$. 

\looseness -1 By definition of the Fourier basis, a $k$-sparse degree $d$ function can be written as a summation of $k$ (Fourier basis) functions, each one depending on at most $d$ input variables. The converse is also true:

\begin{restatable}{proposition}{propDecomposed}
\label{prop:decomposed_function}
Assume $h:\{0, 1\}^n \rightarrow \R$ can be decomposed as follows: $h(x) = \sum\limits_{i=1}^{p} h_i(x_{S_i}), S_i \subseteq [n]$.
That is, each function $h_i: \{0,1\}^{|S_i|} \rightarrow \R$ depends on at most $|S_i|$ variables.  Then,  $h$ is $k=O( \sum\limits_{i=1}^p 2^{|S_i|})$-sparse and of degree $d=\max(|S_1|, \dots, |S_p|)$. (Proof in Appendix~\appref{app:proofs:propositions})
\end{restatable}

The sparsity $k$ and degree $d$ capture a notion of complexity for the underlying function. Intuitively speaking, the sparsity factor $k$ puts a limit on the number of functions in the decomposition, and the degree $d$ puts a limit on the order of interactions among the input variables.

\looseness -1 One can see from Proposition~\ref{prop:decomposed_function}, that modular functions, i.e., functions that can be written as a sum of functions each depending on exactly one variable, are $k=O(n)$-sparse and of degree $d=1$. A slightly more ``complex'' function capturing second-order interactions among the input variables, i.e., a function that can be written as a sum where each term depends on at most two variables is going to be $k=O(n^2)$-sparse and of degree $d=2$. The following proposition generalizes this result.   

\begin{restatable}{proposition}{propDegree}
\label{prop:degree_imlies_sparsity}
Let, $h:\{0,1\}^n \rightarrow \R$ be a pseudo-boolean function and let $d \in \mathbb{N}$ be some constant (w.r.t. $n$). If $h$ is of degree $d$, then, it is $k=O(n^d)$-sparse. (Proof in Appendix~\appref{app:proofs:propositions})
\end{restatable}
This proposition formally shows that limiting the order of interactions among the input variables implies an upper bound on the sparsity.

\vspace*{-1mm}
\section{Many real-world black-box predictors have sparse Fourier transforms}
\vspace*{-1mm}
In this section, we discuss the sparsity of the Fourier transforms of ensembles of trees and why neural networks can be approximated by a sparse Fourier representation because of their spectral bias.  This shows both these classes of functions can be compactly represented in the Fourier basis. The results here will become useful in the next section, where we present our main contribution on how we can leverage this compact representation,  to precisely compute SHAP values cheaply.
\vspace*{-1mm}
\subsection{Spectral bias of fully connected neural networks}
\vspace*{-1mm}
The function a fully connected neural network represents at initialization is a sample from a Gaussian Process (GP) \citep{rasmussen2004gaussian} in the infinite-width limit. Here, the randomness is over the initial weights and biases.  The kernel $K$ of this GP is called the Conjugate Kernel (CK) \citep{daniely2016toward, lee2017deep}.  
As mentioned before, here we investigate the case where the inputs to the neural network are binary, i.e., $\mathcal{X}^d = \{0, 1\}^d$, similar to \citep{yang2019fine, valle2018deep}.
The CK kernel Gram matrix formed on the whole input space $\mathcal{X}^d = \{0, 1\}^d$, has a simple eigenvalue decomposition:
$
\mathcal K \in \R^{2^d \times 2^d}, \mathcal K= \sum\limits_{f\in \{0,1\}^n} \lambda_f u_f u_f^\top, 
$
where $u_f \in \R^{2^d}$ is the eigenvector formed by evaluating the Fourier basis function $\Psi_f$ for different values of $x \in \{0,1\}^n$. Moreover, \citet{yang2019fine} show a weak spectral bias result in terms of the degree of $f$. Namely, the eigenvalues corresponding to higher degree frequencies have smaller values \footnote{To be more precise, they show that the eigenvalues corresponding to even and odd degree frequencies form decreasing sequences. That is, even and odd degrees are considered separately.}. Given the kernel, $K$, and viewing a randomly initialized neural network function evaluated on the boolean $\{0,1\}^n$ as a sample from a GP one can see the following: This sample, roughly speaking, looks like a linear combination of the eigenvectors with the largest eigenvalues. This is due to the fact that a sample of the GP can be obtained as $\sum\limits_{i=1}^{2^n} \lambda_i \bm{w_i} u_i, \bm{w_i} \sim \mathcal{N}(0,1)$. Combining this with the spectral bias results implies that neural networks are low-degree functions when randomly initialized.

Going beyond neural networks at initialization, numerous studies have investigated the behavior of fully connected neural networks trained through (stochastic) gradient descent. \citet{chizat2019lazy, jacot_neural_2018, du2018gradient, allen2019convergence, allen2019convergencernn} found that the weights of infinite-width neural networks after training do not deviate significantly from their initialization, which has been dubbed "lazy training" by \citet{chizat2019lazy}.  \citet{lee2018deep, lee2019wide} showed that training the last layer of an infinite-width randomly initialized neural network for an infinite amount of time corresponds to Gaussian process (GP) posterior inference with a certain kernel. \citet{jacot_neural_2018} extended the aforementioned results to training \emph{all} the layers of a neural network (not just the final layer). They showed the evolution of an infinite-width neural network function can be described by the ``Neural Tangent Kernel'' \citep[\textsc{NTK},][]{jacot2018neural} with the trained network yielding, on average, the posterior mean prediction of the corresponding GP after an infinite amount of training time. \citet{lee2019wide} empirically showed the results carry over to the finite-width setting through extensive experiments.  \citet{yang2019fine} again showed that the $u_f \in \R^{2^d}$ defined above are eigenvectors of the \textsc{NTK} Gram matrix and spectral bias holds. \citet{gorji2023scalable} validated these theoretical findings through extensive experiments in finite-width neural networks by showing that neural networks have ``less tendency'' to learn high-degree frequencies. We refer the reader to Appendix~\appref{subsec:simplicity_bias_review} for a more comprehensive review. 

The aforementioned literature shows that neural networks can be approximated by low-degree functions. By Proposition \ref{prop:degree_imlies_sparsity}, they can be approximated by sparse functions for a large enough sparsity factor $k$. Our experiments in Section~\ref{sec:experiments} provide further evidence that 
neural networks trained on real-world datasets are approximated well with sparse (and therefore compact) Fourier representations.  

\vspace*{-1mm}
\subsection{Sparsity of ensembles of decision trees}
\vspace*{-1mm}
In our context, a decision tree is a rooted binary tree, where each non-leaf node corresponds to one of $n$ binary (zero-one) features, and each leaf node has a real number assigned to it. We denote the function a decision tree represents by $t:\{0, 1\}^n \rightarrow \mathbb{R}$. Let $i \in [n]$ denote the feature corresponding to the root, and let $t_{\mathrm{left}}:\{0,1\}^{n-1}\rightarrow \R$ and $t_{\mathrm{right}}:\{0,1\}^{n-1} \rightarrow \R$ be the left and right sub-trees, respectively. Then the tree can be represented as:
\begin{equation} 
\label{eq:recursive_tree}
t(x) = \frac{1+(-1)^{\langle e_i, x \rangle}}{2} t_{\mathrm{left}}(x) + \frac{1-(-1)^{\langle e_i, x \rangle}}{2} t_{\mathrm{right}}(x)
\end{equation}
where, $e_i \in \{0,1\}^n$ is $i$'th indicator vector.

Thus, the Fourier transform of a decision tree can be computed recursively \citep{kushilevitz1993learning, mansour1994learning}. The degree of a decision tree function of depth $d$ is $d$, and if $|\text{supp}(t_{\mathrm{left}})|=k_{\mathrm{left}}$ and $|\text{supp}(t_{\mathrm{right}})|=k_{\mathrm{right}}$, then $|\text{supp}(t)| \leq 2(k_{\mathrm{left}}+k_{\mathrm{right}})$. As a result, a decision tree function is $k$-sparse, where $k=O(4^d)$, although in some cases, when the decision tree is not balanced or cancellations occur, the Fourier transform can be sparser, i.e., admit a lower $k$, than the above upper bound on the sparsity suggests.

Due to the linearity of the Fourier transform, the Fourier transform of an \emph{ensemble of trees}, such as those produced by the random forest, cat-boost \citep{dorogush2018catboost}, and XGBoost \citep{chen2016xgboost} algorithms/libraries, can be computed by taking the average of the Fourier transform of each tree. If the random forest model has $T$ trees, then its Fourier transform is $k=O(T 4^d)$-sparse and of degree $d$ equal to its maximum depth of the constituent trees. 

\section{Computing SHAP values with Fourier representation of functions}
\vspace*{-1mm}
In the previous section, we saw that many real-world models trained on tabular/discrete data can be exactly represented (in the case of ensembles of decision trees), or efficiently approximated (in the case of neural networks) using a compact (sparse) Fourier representation. We saw neural networks have a tendency to learn approximately low-degree, and hence by Proposition \ref{prop:degree_imlies_sparsity}, sparse functions. This has been attributed in numerous works to the reason why they generalize well and do not overfit despite their over-parameterized nature \citep{valle2018deep, yang2019fine, huh_low-rank_2022, durvasula_there_2020, kalimeris_sgd_2019, neyshabur_exploring_2017, arpit_closer_2017}. We also saw that (ensembles) of decision trees, by nature, have sparse Fourier representations \citep{kushilevitz1993learning, mansour1994learning}. More generally, as made formal in Proposition \ref{prop:decomposed_function} and the remarks after, any ``simple'' function that can be written as a summation of a ``few'' functions each depending on a ``few'' of the input variables is sparse and low-degree in the Fourier domain. 

We propose the following method to approximate SHAP values for black-box functions.  In the first step, given query access to a black-box function, we utilize a sparse Fourier approximation algorithm such as \citep{li2015active, amrollahi2019efficiently, li2015spright} to efficiently extract its sparse and hence compactly represented Fourier approximation. See Appendix~\appref{subsec:sparse_fft_review} for a more detailed explanation. Next, in our second step presented here, we use the Fourier representation to exactly compute SHAP values.


Let $h:\{0,1\}^n \rightarrow \R$ be some predictor that we assume to be $k$-sparse. Assume $[n] \triangleq \{1,\dots, n\}$ as the set of features and let $x^* \in \{0, 1\}^n$ be the instance we are explaining.
Aligned with \citep{lundberg2017unified}, we use
$v_{h}(S) \triangleq \frac{1}{|\mathcal{D}|} \sum\limits_{(x,y) \in \mathcal{D}} h(x^*_S,x_{[n] \setminus S})$ as the value function, which is the average prediction when fixing $x_S^*$ in the (background) dataset $\mathcal{D}$. We compute the Shapley values for a single Fourier basis function $\Psi_f(x)=(-1)^{\langle f, x \rangle}, f \in \{0,1\}^n$ according to Equation~\ref{eqn:shap}:
\vspace*{-1mm}
\begin{align}
\label{eq:shapf1}
\phi_i^{\Psi_f} =
\frac{1}{|\mathcal{D}|}  \sum\limits_{S \subseteq [n] \setminus \{i\}}  \frac{|S|! (n-|S|-1)!}{n!} \cdot 
\sum\limits_{(x,y) \in \mathcal{D}}
\left((-1)^{\langle f, x^*_{S \cup\{i\}} \oplus x_{[n] \setminus S \cup \{i\}} \rangle} - (-1)^{\langle f, x^*_S \oplus x_{[n] \setminus S}\rangle}\right) 
\end{align}

\vspace*{-3mm}
The $\oplus$ operator concatenates two vectors along the same axis. 

This expression still has an exponential sum (in $n$), since we are summing over all subsets $S$. As a main contribution, we find a closed-form analytic expression for the inner summation using a combinatorial argument. This results in the following key Lemma:
\begin{lemma}
\label{lemma:shapf_fourier_basis}
Let $\Psi_f(x)=(-1)^{\langle f, x \rangle}$ be the Fourier basis function for some $f \in \{0,1\}^n$. Then the SHAP value of the Fourier basis function $\Psi_f$ with respect to the background dataset $\mathcal{D}$ is given as:
\begin{equation} 
\label{eq:shapf_fourier_basis}
\phi_i^{\Psi_f}  = -\frac{2f_i}{|\mathcal{D}|} \sum\limits_{(x,y) \in \mathcal{D}}  \mathmybb{1}_{x_i \neq x^*_i}(-1)^{\langle f, x\rangle}  \frac{(|A|+1) \mod 2}{|A|+1}
\end{equation}
where $A \triangleq \{j \in [n] | x_j \neq x^*_j,  j \neq i, f_j=1\}$. (Proof in Appendix~\appref{app:proofs:lemma1})
\end{lemma}

Intuitively, feature $i$ contributes to the prediction only if the Fourier basis function $\psi_f$ is sensitive to the value of feature $i$ i.e. $f_i=1$ and otherwise its contribution is zero. The SHAP values is also dependent on the parity of the number of differences between the query and background samples within support of $f$: many positive and negative terms cancel during the exponential summation. The key combinatorial insight is that we can count the number of cancellations in the exponential sum without computing it and therefore this collapses into a closed-form factor depending on the parity of the ``overlap set'' $A$.

Finally, by the linearity of SHAP values w.r.t. the explained function $h$, and by utilizing Lemma~\ref{lemma:shapf_fourier_basis} we arrive at the final expression for SHAP values of $h$. We present this closed-form expression alongside its computational complexity in our main Theorem:
\begin{theorem}
\label{thm:main}
Let $h:\{0,1\}^n \rightarrow \R$ be a $k$-sparse pseudo-boolean function with Fourier frequencies $f^1, \dots, f^k \in \text{supp}(h)$ and amplitudes $\widehat{h}(f), \forall f \in \text{supp}(f)$. Let $\mathcal{D}$ be a background dataset  of size $|\mathcal{D}|$. Then, Equation~\ref{eq:shapf_final} provides a precise expression for the SHAP value vector $\phi^h=(\phi^h_1,\dots, \phi^h_n)$. One can compute this vector  with $\Theta(n \cdot |\mathcal{D}| \cdot k)$ flops (floating point operations). 
\begin{align}
\label{eq:shapf_final}
\phi_i^{h}  = -\frac{2}{|\mathcal{D}|} \sum\limits_{f \in \text{supp}(h)} \widehat{h}(f) \cdot f_i \cdot
\sum\limits_{(x,y) \in \mathcal{D}}  \mathmybb{1}_{x_i \neq x^*_i}(-1)^{\langle f, x\rangle}  \frac{(|A|+1) \mod 2}{|A|+1} 
\end{align}
where $A$ is the same as in Lemma~\ref{lemma:shapf_fourier_basis}. (Proof in Appendix~\appref{app:proofs:thm2})
\end{theorem}

\vspace*{-1mm}

Theorem~\ref{thm:main} provides a computationally efficient method for deriving SHAP values from a Fourier representation of a function $h$. The resulting SHAP values are {\em exact}, that is, if the Fourier representation is precise, then the SHAP values are also exact. This contrasts with \textsc{KernelSHAP}, which approximates SHAP values via stochastic sampling, requiring convergence checks to ensure accuracy. In our method, any approximation arises solely from estimating a sparse Fourier representation, and only in the black-box setting.

More importantly, the sum in Equation~\ref {eq:shapf_final} is {\em tractable}.  This is because we overcome the exponential sum involved in Equation~\ref{eqn:shap} by analytically computing the sum, with a combinatorial argument, for a single Fourier basis function. We show that the number of flops that are required to compute SHAP values in Theorem~\ref{thm:main} is asymptotically equal to $\Theta(n \cdot |\mathcal{D}| \cdot k)$.
We note that the $|\mathcal{D}|$ and $k$ factors in the asymptotic computational complexity arise from the two summations present in Equation~\ref {eq:shapf_final}. Through this expression, we are able to ``linearize'' the computation of SHAP values to a summation over the Fourier coefficients and the background data set. This allows us to maximize parallelization on multiple cores and/or GPUs to speed up the computation significantly. Therefore, in the presence of multiple cores or a GPU, we can get a speedup equal to the level of parallelization, as each core or worker can compute one part of this summation.

We implement our algorithm called \textsc{FourierShap} using JAX \citep{jax2018github}, which allows fast vectorized operations on GPUs. Each term within the summations of Equation~\ref{eq:shapf_final} can be implemented with simple vector operations. Furthermore, summations over the $k$ different frequencies in the support of $h$ and also background data points can both be efficiently implemented and parallelized using this library using its \textsc{vmap} operator. We perform all upcoming experiments on a single GPU.
Nevertheless, we believe faster computation can also be achieved by crafting dedicated code designed to efficiently compute Equation~\ref {eq:shapf_final} on GPU.

\looseness -1 Finally, we note that the function approximation (first step) is only done once, i.e., we compute the sparse Fourier approximation of the black-box only once. This is typically the most expensive part of the computation. For any new query to be explained, we resort to an efficient implementation of Equation~\ref{eq:shapf_final}. As our experiments will show, this yields orders of magnitudes faster computation than previous methods such as \textsc{KernelShap} where, as mentioned before in Section \ref{sec:background}, each explained instance requires solving an expensive optimization problem.

\vspace*{-2mm}
\section{Experiments}
\vspace*{-3mm}
\label{sec:experiments}
We assess the performance of our algorithm, \textsc{FourierSHAP}, on four different real-world datasets of varying nature and dimensionality. Three of our datasets are  related to protein fitness landscapes \citep{poelwijk_learning_2019, wu_adaptation_2016, sarkisyan_local_2016} and are referred to as ``Entacmaea'' (dimension $n=13$), ``GB1'' ($n=80$), and ``avGFP'' ($n=236$) respectively.  The fourth dataset is a GPU-tuning \citep{nugteren_cltune_2015} dataset referred to as ``SGEMM'' ($n=40$). The features of these datasets are binary (zero-one) and/or categorical with standard one-hot encodings. See Appendix~\appref{app:sec:datasets} for dataset details. 

For the Entacmaea and SGEMM datasets, we train fully connected neural networks  with 3 hidden layers containing 300 neurons each. For GB1 we train ensembles of trees models of varying depths using the random forest algorithm and for avGFP we train again, ensembles of trees models of varying depths using the cat-boost algorithm/library \citep{dorogush2018catboost}.

\begin{figure*}[!tb]
\vspace*{-5mm}
    \centering
    \includegraphics[width=0.9\linewidth]{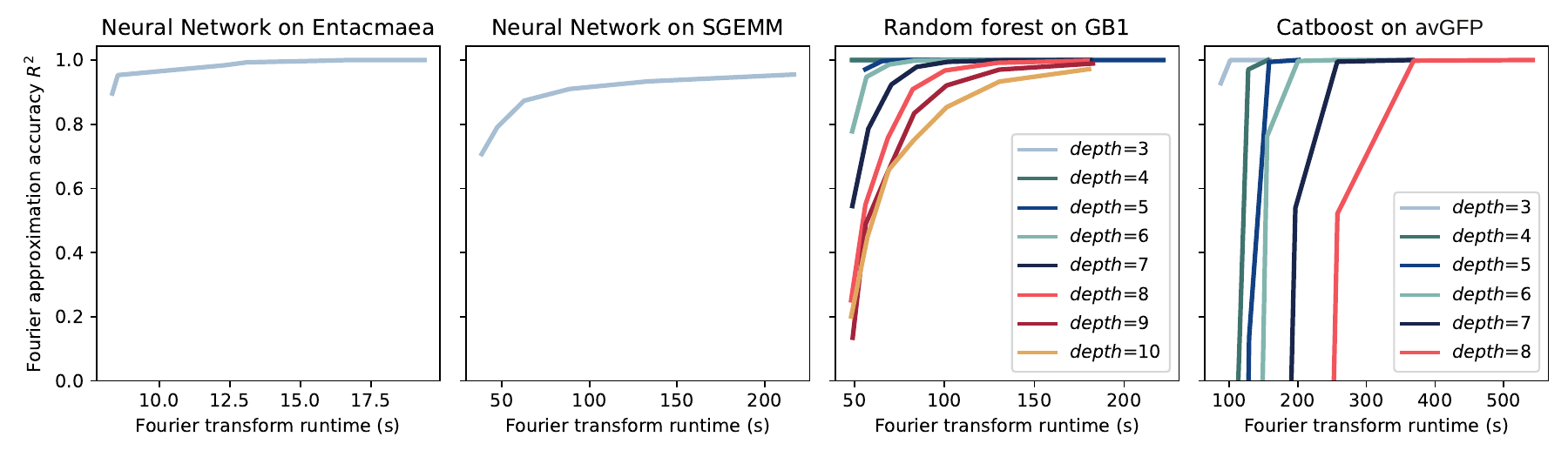}
     \vspace*{-2mm}
    \caption{Step 1 of \textsc{FourierShap}: Accuracy of the Fourier transform (in approximating the black-box function) vs. runtime of the sparse Fourier algorithm. The accuracy is evaluated by $R^2$ score and comparing the outputs of the black-box and the Fourier representation on a uniformly generated random dataset on the Boolean cube $\{0,1\}^n$. For a fixed level of accuracy, higher depth trees require a higher number of Fourier coefficients $k$ therefore a higher runtime. For the case of trees, we eventually are able to reach a perfect approximation since the underlying function is truly sparse. }
\label{fig:fourier_runtime}
\end{figure*}

\textbf{Black-box setting.} \looseness -1 The first step of the \textsc{FourierSHAP} algorithm is to compute a sparse Fourier approximation of the black-box model. We use a GPU implementation of a sparse Walsh-Hadamard Transform (sparse WHT) a.k.a Fourier transform algorithm \citep{amrollahi2019efficiently} for each of the four trained models. The algorithm accepts a sparsity parameter $k$ which is the sparsity of the computed Fourier representation. Higher sparsity parameter $k$ results in a better function approximation but the sparse-WHT runtime increase linearly in $k$ as well. In Figure~\ref{fig:fourier_runtime} we plot the accuracy of the Fourier function approximation as measured by the $R^2$-score for different values of $k$ (which result in different runtimes). The $R^2$ score is computed over a dataset formed by randomly sampling the Boolean cube $\{0, 1\}^n$. 

The second step of \textsc{FourierSHAP} utilizes the Fourier approximation to compute \textsc{SHAP} values using Equation \ref{eq:shapf_final}. We implement this step using JAX library~\citep{jax2018github}, and run it on a GPU. For each model to be explained, we choose four different values for the number of background samples and four different values for the number of query points to be explained, resulting in a total of 16 runs of \textsc{FourierSHAP} for each model. Error bars capture these variations. We take the runtime of \textsc{KernelSHAP}, with Github repo default settings, to be the base runtime all other methods are compared to, i.e., we assume its runtime is 1 unit. 

\begin{figure*}[!t]
    \vspace*{-3mm}
    \centering
     \begin{subfigure}[b]{0.44\linewidth}
         \centering
         \includegraphics[width=\linewidth]{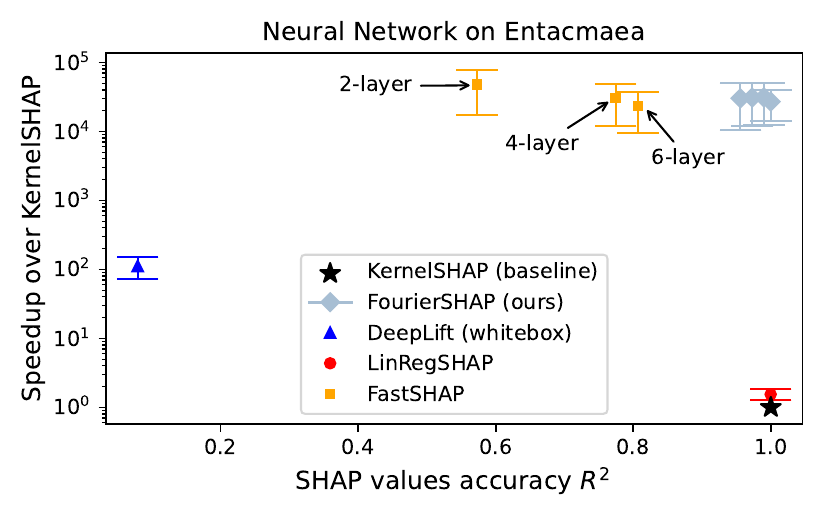}
         \label{5}
     \end{subfigure}
     \begin{subfigure}[b]{0.44\linewidth}
         \centering
         \includegraphics[width=\linewidth]{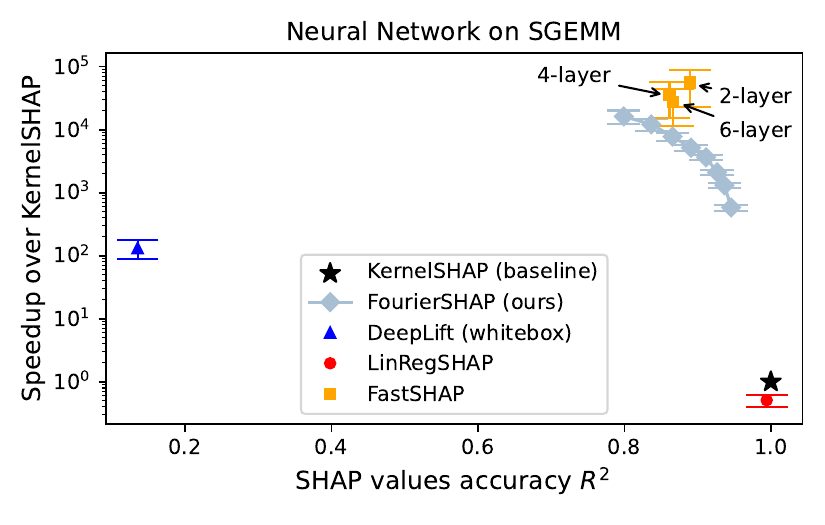}
         \label{6}
     \end{subfigure}
     \\ \vspace{-0.5cm}
    \begin{subfigure}[b]{0.45\linewidth}
         \centering
         \includegraphics[width=\linewidth]{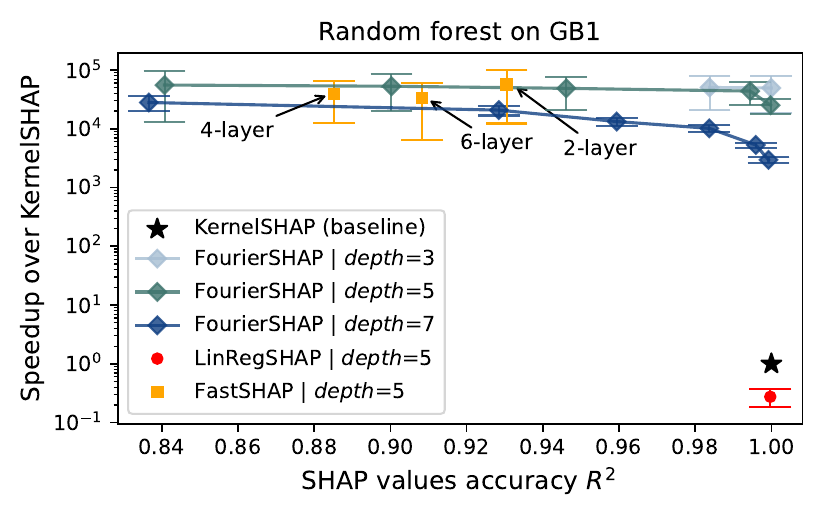}
         \label{7}
     \end{subfigure}
    \begin{subfigure}[b]{0.45\linewidth}
         \centering
         \includegraphics[width=\linewidth]{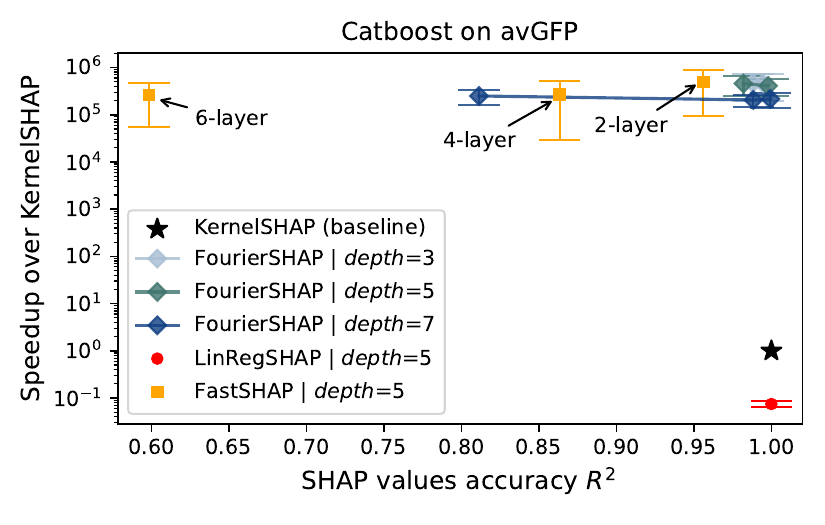}
         \label{8}
     \end{subfigure}
     \vspace*{-6mm}
    \caption{Speedup vs. Accuracy. Speedup of different algorithms is reported as a multiple compared to the runtime of \textsc{KernelShap}. Accuracy is quantified by the $R^2$-score against ground truth SHAP values. \textbf{\textsc{DeepLift}} is a white-box algorithm for neural networks. \textbf{\textsc{LinRegShap}} is a black-box algorithm and a variance-reduced version of \textsc{KernelShap}. \textbf{\textsc{FastShap}}, a black-box algorithm, is a trained MLP to output SHAP values given inputs in one forward pass. We test MLPs of three different sizes for \textbf{\textsc{FastShap}}. \textbf{\textsc{FourierShap}} is ours. We are 10-10000x faster than \textsc{LinRegShap} on all dataset/model variations. More notably, we outperform DeepLift in the neural network model setting even though we assume only query access (black-box setting). We achieve higher accuracy than \textsc{FastShap} in 3/4 settings, while enabling a fine-grained control over the speed-accuracy trade-off.}
\label{fig:speedup}
\vspace*{-2mm}
\end{figure*}

In order to measure the accuracy of the SHAP values produced by our and other algorithms we need ground truth SHAP values. We use \textsc{KernelSHAP} to this extent.   Note that \textsc{KernelSHAP} is inherently an \emph{approximation} method for computing interventional SHAP values. However, this approximation becomes more precise by sampling more coalition subsets. Therefore, to generate ground truth values, for each dataset, we sample more and more coalition subsets and check for convergence in these values. A more detailed explanation can be found in Appendix~\ref{app:kernel_shap_is_ground_truth}.

We compute the $R^2$ values of Shapley values computed by \textsc{FourierSHAP} (ours) vs. ground truth values to evaluate accuracy.  For our method, a higher sparsity $k$ for the Fourier representation results in a more accurate function approximation therefore higher $R^2$ values for the SHAP value quality. On the other hand, a higher $k$ results in a slower runtime as Equation~\ref{eq:shapf_final} is  a sum over the $k$ different frequencies. In Figure~\ref{fig:speedup} we plot this trade-off.

We compare against the following baselines in Figure \ref{fig:speedup}.  The first is \textsc{LinRegShap}, a variance-reduced version of \textsc{KernelSHAP} \citep{covert2020improving}. We found that although this algorithm requires fewer queries from the black-box, it takes orders of magnitudes longer to run compared to ours. Secondly, for the neural network models, we also compare against a state-of-the-art {\em white-box} method -- \textsc{DeepLift} \citep{shrikumar2017learning}. This algorithm, 
requires access to the neural network's activations in all layers. In comparison, we achieve a 10-100x speedup 
while being both more accurate and only assuming query access to the neural net (true black-box setting). 

Finally, we compare against \textsc{FastShap} \citep{jethani2021fastshap}, which is the most comparable to our setting, i.e., model-agnostic with amortized cost. We provide a higher accuracy for similar speedup values compared to \textsc{FastShap} in 3/4 settings. More importantly, we can see that by controlling the sparsity parameter $k$ of the Fourier function approximation, we can control the tradeoff between accuracy and speed in a reliable and fine-grained manner. Whereas for \textsc{FastShap}, the accuracy of the SHAP values relies on the functional approximation properties of the MLP which directly produce SHAP values. As seen in Figure \ref{fig:speedup}, the MLP can behave in unpredictable ways.  In this figure, we can see that increasing the depth of the model (and hence the approximation capability of the MLP) does not have a reliable effect on SHAP accuracy.  Finally, in \textsc{FastShap}, for a different choice of background dataset, a new MLP model has to be trained from scratch since the MLP approximates the process of directly computing SHAP values. In contrast, we can support different sets of background datasets since our Fourier functional approximation is performed on the black-box predictor, and not on the whole process of producing SHAP values end-to-end (\textsc{FastShap}). 

\textbf{Tree setting.}
\textsc{FourierSHAP} can also be utilized for the computation of \textsc{SHAP} values for (ensembles of) trees in the white-box setting, i.e., where full access to the tree's structure is available. In this setting, the exact sparse Fourier representation of an ensemble of trees can be efficiently computed (the first step of \textsc{FourierSHAP}) using Equation~\ref{eq:recursive_tree}. 
With the exact Fourier representation at hand, \textsc{SHAP} values can be {\em efficiently and exactly} computed using Equation~\ref{eq:shapf_final}, the second step of \textsc{FourierSHAP}.
This makes our method a highly parallelizable alternative for \textsc{TreeShap} \citep{lundberg2020local} with the potential for order of magnitude speedups.

\begin{figure*}[!t]
\vspace*{-5mm}
    \centering
    \includegraphics[width=0.9\linewidth]{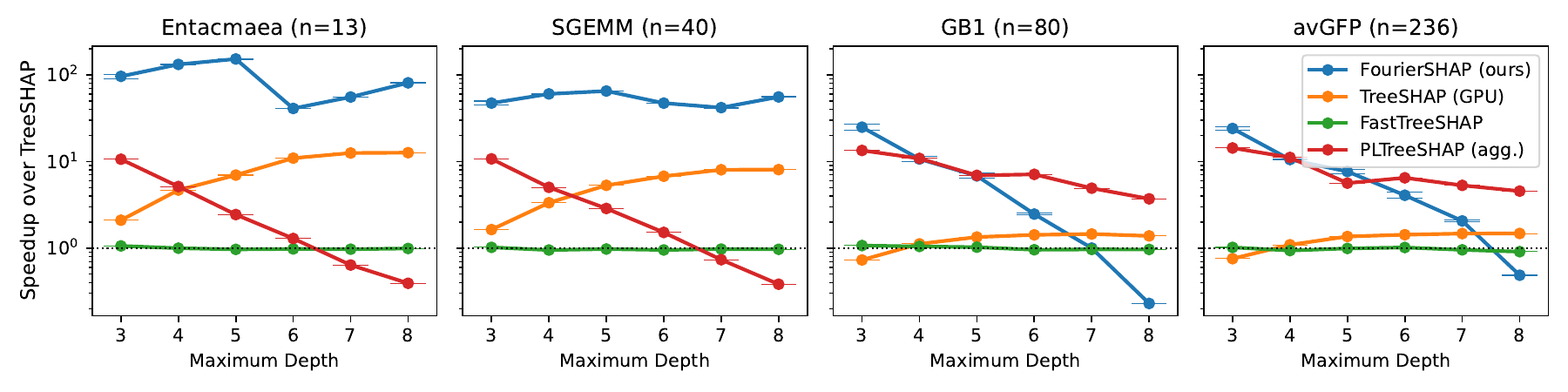}
    \vspace{-2mm}
    \caption{Speedup vs. depth of tree, for different algorithms, reported as a multiple compared to the runtime of \textsc{TreeSHAP}. \textbf{\textsc{FourierSHAP}} is ours. As other baselines we have a GPU implementation of \textbf{\textsc{TreeSHAP}}, \textbf{\textsc{FastTreeSHAP}}, and \textbf{\textsc{PLTreeSHAP}}. We achieve order of magnitude speedups over all depths on the Entacmaea and SGEMM datasets. We also achieve significant speedups in the other two datasets; however, the edge diminishes as the maximum depth increases.}
\label{fig:whitebox}
\vspace*{-4mm}
\end{figure*}

To demonstrate our method's ability in fast computation of SHAP values in this setting, we compute SHAP values for random forests fitted on all aforementioned real-world datasets.
We compare \textsc{FourierSHAP}'s speedup over \textsc{TreeSHAP}~\citep{lundberg2020local}, to \textsc{FastTreeSHAP}~\citep{yang2021fast}, a fast implementation of \textsc{TreeSHAP}, the GPU implementation of \textsc{TreeSHAP}~\citep{mitchell2022gputreeshap} and \textsc{PLTreeSHAP}~\citep{zern2023interventional}.
To the best of our knowledge, these are the fastest available frameworks for computation of the ``interventional'' \textsc{SHAP} values. Figure~\ref{fig:whitebox} shows the superior speed of our method over all state-of-the-art algorithms in most settings, which weakens with increased depth and number of features, resulting in a higher count of frequencies and greater computational overhead. Note that the SHAP values computed by all methods are precise and identical to the values produced by \textsc{TreeSHAP}, which is to be expected for \textsc{FourierSHAP}, given access to exact Fourier representation of random forests.


\section*{Conclusions}
\vspace*{-2mm}
We illustrated in theory and practice how many black-box functions  can be represented or efficiently approximated by a compact Fourier representation. We proved that SHAP values of Fourier basis functions admit a closed form expression, and therefore, we can compute SHAP values efficiently using the compact Fourier representation. Moreover, this closed form expression is amenable to parallelization. These two factors helped us in gaining speedups of 10-10000x over baseline methods for the computation of SHAP values. We discuss the limitations \& future works in Appendix~\ref{app:limitations}.

\section*{Acknowledgments}
This work was supported by the Swiss National Science Foundation (SNSF) through National Centre of Competence in Research (NCCR) Catalysis (Grant No.\ 180544). We thank Bhavya Sukhija for valuable feedback and help during the review and rebuttal process.

\medskip
\clearpage
\bibliography{bibs}

\begin{thebibliography}{73}
\providecommand{\natexlab}[1]{#1}
\providecommand{\url}[1]{\texttt{#1}}
\expandafter\ifx\csname urlstyle\endcsname\relax
  \providecommand{\doi}[1]{doi: #1}\else
  \providecommand{\doi}{doi: \begingroup \urlstyle{rm}\Url}\fi

\bibitem[Aas et~al.(2021)Aas, Jullum, and L{\o}land]{aas2021explaining}
Kjersti Aas, Martin Jullum, and Anders L{\o}land.
\newblock Explaining individual predictions when features are dependent: More accurate approximations to shapley values.
\newblock \emph{Artificial Intelligence}, 298:\penalty0 103502, 2021.

\bibitem[Allen-Zhu et~al.(2019{\natexlab{a}})Allen-Zhu, Li, and Song]{allen2019convergence}
Zeyuan Allen-Zhu, Yuanzhi Li, and Zhao Song.
\newblock A convergence theory for deep learning via over-parameterization.
\newblock In \emph{International Conference on Machine Learning}, pages 242--252. PMLR, 2019{\natexlab{a}}.

\bibitem[Allen-Zhu et~al.(2019{\natexlab{b}})Allen-Zhu, Li, and Song]{allen2019convergencernn}
Zeyuan Allen-Zhu, Yuanzhi Li, and Zhao Song.
\newblock On the convergence rate of training recurrent neural networks.
\newblock \emph{Advances in neural information processing systems}, 32, 2019{\natexlab{b}}.

\bibitem[Amrollahi et~al.(2019)Amrollahi, Zandieh, Kapralov, and Krause]{amrollahi2019efficiently}
Andisheh Amrollahi, Amir Zandieh, Michael Kapralov, and Andreas Krause.
\newblock Efficiently learning fourier sparse set functions.
\newblock In \emph{Advances in Neural Information Processing Systems}, pages 15120--15129, 2019.

\bibitem[Ancona et~al.(2018)Ancona, Ceolini, {\"O}ztireli, and Gross]{ancona2018towards}
Marco Ancona, Enea Ceolini, Cengiz {\"O}ztireli, and Markus Gross.
\newblock Towards better understanding of gradient-based attribution methods for deep neural networks.
\newblock In \emph{International Conference on Learning Representations}, 2018.

\bibitem[Arenas et~al.(2021)Arenas, Barcel{\'o}, Bertossi, and Monet]{arenas2021tractability}
Marcelo Arenas, Pablo Barcel{\'o}, Leopoldo Bertossi, and Mika{\"e}l Monet.
\newblock The tractability of shap-score-based explanations for classification over deterministic and decomposable boolean circuits.
\newblock In \emph{Proceedings of the AAAI Conference on Artificial Intelligence}, volume~35, pages 6670--6678, 2021.

\bibitem[Arpit et~al.(2017)Arpit, Jastrzebski, Ballas, Krueger, Bengio, Kanwal, Maharaj, Fischer, Courville, Bengio, and Lacoste-Julien]{arpit_closer_2017}
Devansh Arpit, Stanisław Jastrzebski, Nicolas Ballas, David Krueger, Emmanuel Bengio, Maxinder~S. Kanwal, Tegan Maharaj, Asja Fischer, Aaron Courville, Yoshua Bengio, and Simon Lacoste-Julien.
\newblock A {Closer} {Look} at {Memorization} in {Deep} {Networks}.
\newblock In \emph{Proceedings of the 34th {International} {Conference} on {Machine} {Learning}}, pages 233--242. PMLR, July 2017.
\newblock ISSN: 2640-3498.

\bibitem[Basri et~al.(2020)Basri, Galun, Geifman, Jacobs, Kasten, and Kritchman]{basri2020frequency}
Ronen Basri, Meirav Galun, Amnon Geifman, David Jacobs, Yoni Kasten, and Shira Kritchman.
\newblock Frequency bias in neural networks for input of non-uniform density.
\newblock In \emph{International Conference on Machine Learning}, pages 685--694. PMLR, 2020.

\bibitem[Bifet et~al.(2022)Bifet, Read, Xu, et~al.]{bifet2022linear}
Albert Bifet, Jesse Read, Chao Xu, et~al.
\newblock Linear tree shap.
\newblock \emph{Advances in Neural Information Processing Systems}, 35:\penalty0 25818--25828, 2022.

\bibitem[Bradbury et~al.(2018)Bradbury, Frostig, Hawkins, Johnson, Leary, Maclaurin, Necula, Paszke, Vander{P}las, Wanderman-{M}ilne, and Zhang]{jax2018github}
James Bradbury, Roy Frostig, Peter Hawkins, Matthew~James Johnson, Chris Leary, Dougal Maclaurin, George Necula, Adam Paszke, Jake Vander{P}las, Skye Wanderman-{M}ilne, and Qiao Zhang.
\newblock {JAX}: composable transformations of {P}ython+{N}um{P}y programs, 2018.
\newblock URL \url{http://github.com/google/jax}.

\bibitem[Cao et~al.(2019)Cao, Fang, Wu, Zhou, and Gu]{cao2019towards}
Yuan Cao, Zhiying Fang, Yue Wu, Ding-Xuan Zhou, and Quanquan Gu.
\newblock Towards understanding the spectral bias of deep learning.
\newblock \emph{arXiv preprint arXiv:1912.01198}, 2019.

\bibitem[Chen et~al.(2020)Chen, Janizek, Lundberg, and Lee]{chen2020true}
Hugh Chen, Joseph~D Janizek, Scott Lundberg, and Su-In Lee.
\newblock True to the model or true to the data?
\newblock \emph{arXiv preprint arXiv:2006.16234}, 2020.

\bibitem[Chen and Guestrin(2016)]{chen2016xgboost}
Tianqi Chen and Carlos Guestrin.
\newblock Xgboost: A scalable tree boosting system.
\newblock In \emph{Proceedings of the 22nd acm sigkdd international conference on knowledge discovery and data mining}, pages 785--794, 2016.

\bibitem[Cheraghchi and Indyk(2017)]{cheraghchi2017nearly}
Mahdi Cheraghchi and Piotr Indyk.
\newblock Nearly optimal deterministic algorithm for sparse walsh-hadamard transform.
\newblock \emph{ACM Transactions on Algorithms (TALG)}, 13\penalty0 (3):\penalty0 1--36, 2017.

\bibitem[Chizat et~al.(2019)Chizat, Oyallon, and Bach]{chizat2019lazy}
Lenaic Chizat, Edouard Oyallon, and Francis Bach.
\newblock On lazy training in differentiable programming.
\newblock \emph{Advances in Neural Information Processing Systems}, 32, 2019.

\bibitem[Cho and Saul(2009)]{cho2009kernel}
Youngmin Cho and Lawrence Saul.
\newblock Kernel methods for deep learning.
\newblock \emph{Advances in neural information processing systems}, 22, 2009.

\bibitem[Covert and Lee(2020)]{covert2020improving}
Ian Covert and Su-In Lee.
\newblock Improving kernelshap: Practical shapley value estimation via linear regression.
\newblock \emph{arXiv preprint arXiv:2012.01536}, 2020.

\bibitem[Daniely(2017)]{daniely2017sgd}
Amit Daniely.
\newblock Sgd learns the conjugate kernel class of the network.
\newblock \emph{Advances in Neural Information Processing Systems}, 30, 2017.

\bibitem[Daniely et~al.(2016)Daniely, Frostig, and Singer]{daniely2016toward}
Amit Daniely, Roy Frostig, and Yoram Singer.
\newblock Toward deeper understanding of neural networks: The power of initialization and a dual view on expressivity.
\newblock \emph{Advances in neural information processing systems}, 29, 2016.

\bibitem[Datta et~al.(2016)Datta, Sen, and Zick]{datta2016algorithmic}
Anupam Datta, Shayak Sen, and Yair Zick.
\newblock Algorithmic transparency via quantitative input influence: Theory and experiments with learning systems.
\newblock In \emph{2016 IEEE symposium on security and privacy (SP)}, pages 598--617. IEEE, 2016.

\bibitem[Dorogush et~al.(2018)Dorogush, Ershov, and Gulin]{dorogush2018catboost}
Anna~Veronika Dorogush, Vasily Ershov, and Andrey Gulin.
\newblock Catboost: gradient boosting with categorical features support.
\newblock \emph{arXiv preprint arXiv:1810.11363}, 2018.

\bibitem[Du et~al.(2018)Du, Zhai, Poczos, and Singh]{du2018gradient}
Simon~S Du, Xiyu Zhai, Barnabas Poczos, and Aarti Singh.
\newblock Gradient descent provably optimizes over-parameterized neural networks.
\newblock \emph{arXiv preprint arXiv:1810.02054}, 2018.

\bibitem[Durvasula and Liter(2020)]{durvasula_there_2020}
Karthik Durvasula and Adam Liter.
\newblock There is a simplicity bias when generalising from ambiguous data.
\newblock \emph{Phonology}, 37\penalty0 (2):\penalty0 177--213, May 2020.
\newblock ISSN 0952-6757, 1469-8188.
\newblock \doi{10.1017/S0952675720000093}.
\newblock Publisher: Cambridge University Press.

\bibitem[Fan and Wang(2020)]{fan2020spectra}
Zhou Fan and Zhichao Wang.
\newblock Spectra of the conjugate kernel and neural tangent kernel for linear-width neural networks.
\newblock \emph{Advances in neural information processing systems}, 33:\penalty0 7710--7721, 2020.

\bibitem[Gorji et~al.(2023)Gorji, Amrollahi, and Krause]{gorji2023scalable}
Ali Gorji, Andisheh Amrollahi, and Andreas Krause.
\newblock A scalable walsh-hadamard regularizer to overcome the low-degree spectral bias of neural networks.
\newblock In \emph{Uncertainty in Artificial Intelligence}, pages 723--733. PMLR, 2023.

\bibitem[Gromping(2007)]{gromping2007estimators}
Ulrike Gromping.
\newblock Estimators of relative importance in linear regression based on variance decomposition.
\newblock \emph{The American Statistician}, 61\penalty0 (2):\penalty0 139--147, 2007.

\bibitem[Hazan and Jaakkola(2015)]{hazan2015steps}
Tamir Hazan and Tommi Jaakkola.
\newblock Steps toward deep kernel methods from infinite neural networks.
\newblock \emph{arXiv e-prints}, pages arXiv--1508, 2015.

\bibitem[Huang et~al.(2024)Huang, Chen, and Lee]{huang2024updates}
Y.~Huang, Y.~Chen, and J.~Lee.
\newblock Updates on the complexity of shap scores.
\newblock In \emph{Proceedings of the 33rd International Joint Conference on Artificial Intelligence (IJCAI)}, 2024.

\bibitem[Huh et~al.(2022)Huh, Mobahi, Zhang, Cheung, Agrawal, and Isola]{huh_low-rank_2022}
Minyoung Huh, Hossein Mobahi, Richard Zhang, Brian Cheung, Pulkit Agrawal, and Phillip Isola.
\newblock The {Low}-{Rank} {Simplicity} {Bias} in {Deep} {Networks}, April 2022.
\newblock arXiv:2103.10427 [cs].

\bibitem[Jacot et~al.(2018{\natexlab{a}})Jacot, Gabriel, and Hongler]{jacot2018neural}
Arthur Jacot, Franck Gabriel, and Cl{\'e}ment Hongler.
\newblock Neural tangent kernel: Convergence and generalization in neural networks.
\newblock \emph{Advances in neural information processing systems}, 31, 2018{\natexlab{a}}.

\bibitem[Jacot et~al.(2018{\natexlab{b}})Jacot, Gabriel, and Hongler]{jacot_neural_2018}
Arthur Jacot, Franck Gabriel, and Clement Hongler.
\newblock Neural {Tangent} {Kernel}: {Convergence} and {Generalization} in {Neural} {Networks}.
\newblock In \emph{Advances in {Neural} {Information} {Processing} {Systems}}, volume~31. Curran Associates, Inc., 2018{\natexlab{b}}.

\bibitem[Janzing et~al.(2020)Janzing, Minorics, and Bloebaum]{pmlr-v108-janzing20a}
Dominik Janzing, Lenon Minorics, and Patrick Bloebaum.
\newblock Feature relevance quantification in explainable ai: A causal problem.
\newblock In Silvia Chiappa and Roberto Calandra, editors, \emph{Proceedings of the Twenty Third International Conference on Artificial Intelligence and Statistics}, volume 108 of \emph{Proceedings of Machine Learning Research}, pages 2907--2916. PMLR, 26--28 Aug 2020.
\newblock URL \url{https://proceedings.mlr.press/v108/janzing20a.html}.

\bibitem[Jethani et~al.(2021)Jethani, Sudarshan, Covert, Lee, and Ranganath]{jethani2021fastshap}
Neil Jethani, Mukund Sudarshan, Ian~Connick Covert, Su-In Lee, and Rajesh Ranganath.
\newblock Fastshap: Real-time shapley value estimation.
\newblock In \emph{International Conference on Learning Representations}, 2021.

\bibitem[Kalimeris et~al.(2019)Kalimeris, Kaplun, Nakkiran, Edelman, Yang, Barak, and Zhang]{kalimeris_sgd_2019}
Dimitris Kalimeris, Gal Kaplun, Preetum Nakkiran, Benjamin Edelman, Tristan Yang, Boaz Barak, and Haofeng Zhang.
\newblock {SGD} on {Neural} {Networks} {Learns} {Functions} of {Increasing} {Complexity}.
\newblock In \emph{Advances in {Neural} {Information} {Processing} {Systems}}, volume~32. Curran Associates, Inc., 2019.

\bibitem[Kushilevitz and Mansour(1993)]{kushilevitz1993learning}
Eyal Kushilevitz and Yishay Mansour.
\newblock Learning decision trees using the fourier spectrum.
\newblock \emph{SIAM Journal on Computing}, 22:\penalty0 1331--1348, 1993.

\bibitem[Kwon et~al.(2021)Kwon, Rivas, and Zou]{kwon2021efficient}
Yongchan Kwon, Manuel~A Rivas, and James Zou.
\newblock Efficient computation and analysis of distributional shapley values.
\newblock In \emph{International Conference on Artificial Intelligence and Statistics}, pages 793--801. PMLR, 2021.

\bibitem[Lee et~al.(2017)Lee, Bahri, Novak, Schoenholz, Pennington, and Sohl-Dickstein]{lee2017deep}
Jaehoon Lee, Yasaman Bahri, Roman Novak, Samuel~S Schoenholz, Jeffrey Pennington, and Jascha Sohl-Dickstein.
\newblock Deep neural networks as gaussian processes.
\newblock \emph{arXiv preprint arXiv:1711.00165}, 2017.

\bibitem[Lee et~al.(2018)Lee, Bahri, Novak, Schoenholz, Pennington, and Sohl-Dickstein]{lee2018deep}
Jaehoon Lee, Yasaman Bahri, Roman Novak, Samuel~S Schoenholz, Jeffrey Pennington, and Jascha Sohl-Dickstein.
\newblock Deep neural networks as gaussian processes.
\newblock In \emph{International Conference on Learning Representations}, 2018.

\bibitem[Lee et~al.(2019)Lee, Xiao, Schoenholz, Bahri, Novak, Sohl-Dickstein, and Pennington]{lee2019wide}
Jaehoon Lee, Lechao Xiao, Samuel Schoenholz, Yasaman Bahri, Roman Novak, Jascha Sohl-Dickstein, and Jeffrey Pennington.
\newblock Wide neural networks of any depth evolve as linear models under gradient descent.
\newblock \emph{Advances in neural information processing systems}, 32, 2019.

\bibitem[Li and Ramchandran(2015)]{li2015active}
Xiao Li and Kannan Ramchandran.
\newblock An active learning framework using sparse-graph codes for sparse polynomials and graph sketching.
\newblock \emph{Advances in Neural Information Processing Systems}, 28, 2015.

\bibitem[Li et~al.(2015)Li, Bradley, Pawar, and Ramchandran]{li2015spright}
Xiao Li, Joseph~K Bradley, Sameer Pawar, and Kannan Ramchandran.
\newblock Spright: A fast and robust framework for sparse walsh-hadamard transform.
\newblock \emph{arXiv preprint arXiv:1508.06336}, 2015.

\bibitem[Lundberg and Lee(2017)]{lundberg2017unified}
Scott~M Lundberg and Su-In Lee.
\newblock A unified approach to interpreting model predictions.
\newblock In \emph{Advances in Neural Information Processing Systems}, volume~30, 2017.
\newblock URL \url{https://proceedings.neurips.cc/paper/2017/file/8a20a8621978632d76c43dfd28b67767-Paper.pdf}.

\bibitem[Lundberg et~al.(2020)Lundberg, Erion, Chen, DeGrave, Prutkin, Nair, Katz, Himmelfarb, Bansal, and Lee]{lundberg2020local}
Scott~M Lundberg, Gabriel Erion, Hugh Chen, Alex DeGrave, Jordan~M Prutkin, Bala Nair, Ronit Katz, Jonathan Himmelfarb, Nisha Bansal, and Su-In Lee.
\newblock From local explanations to global understanding with explainable ai for trees.
\newblock \emph{Nature machine intelligence}, 2\penalty0 (1):\penalty0 56--67, 2020.

\bibitem[Mansour(1994)]{mansour1994learning}
Yishay Mansour.
\newblock Learning boolean functions via the fourier transform.
\newblock In \emph{Theoretical advances in neural computation and learning}, pages 391--424. Springer, 1994.

\bibitem[Marzouk et~al.(2024)Marzouk, Belle, and Van~den Broeck]{marzouk2024markovian}
A.~Marzouk, V.~Belle, and G.~Van~den Broeck.
\newblock On the tractability of shap explanations under markovian distributions.
\newblock In \emph{Proceedings of the 41st International Conference on Machine Learning (ICML)}, 2024.

\bibitem[Marzouk et~al.(2025)Marzouk, Belle, and Van~den Broeck]{marzouk2025many}
A.~Marzouk, V.~Belle, and G.~Van~den Broeck.
\newblock On the computational tractability of the (many) shapley values.
\newblock In \emph{Proceedings of the 28th International Conference on Artificial Intelligence and Statistics (AISTATS)}, 2025.

\bibitem[Mitchell et~al.(2022{\natexlab{a}})Mitchell, Cooper, Frank, and Holmes]{mitchell2022sampling}
Rory Mitchell, Joshua Cooper, Eibe Frank, and Geoffrey Holmes.
\newblock Sampling permutations for shapley value estimation.
\newblock \emph{Journal of Machine Learning Research}, 23\penalty0 (43):\penalty0 1--46, 2022{\natexlab{a}}.

\bibitem[Mitchell et~al.(2022{\natexlab{b}})Mitchell, Frank, and Holmes]{mitchell2022gputreeshap}
Rory Mitchell, Eibe Frank, and Geoffrey Holmes.
\newblock Gputreeshap: Massively parallel exact calculation of shap scores for tree ensembles, 2022{\natexlab{b}}.

\bibitem[Neal(1996)]{neal1996priors}
Radford~M Neal.
\newblock Priors for infinite networks.
\newblock In \emph{Bayesian Learning for Neural Networks}, pages 29--53. Springer, 1996.

\bibitem[Neyshabur et~al.(2017)Neyshabur, Bhojanapalli, McAllester, and Srebro]{neyshabur_exploring_2017}
Behnam Neyshabur, Srinadh Bhojanapalli, David McAllester, and Nathan Srebro.
\newblock Exploring {Generalization} in {Deep} {Learning}, July 2017.
\newblock arXiv:1706.08947 [cs].

\bibitem[Nugteren and Codreanu(2015)]{nugteren_cltune_2015}
Cedric Nugteren and Valeriu Codreanu.
\newblock {CLTune}: {A} {Generic} {Auto}-{Tuner} for {OpenCL} {Kernels}.
\newblock In \emph{2015 {IEEE} 9th {International} {Symposium} on {Embedded} {Multicore}/{Many}-core {Systems}-on-{Chip}}, pages 195--202, September 2015.
\newblock \doi{10.1109/MCSoC.2015.10}.

\bibitem[Owen(2014)]{owen2014sobol}
Art~B Owen.
\newblock Sobol indices and shapley value.
\newblock \emph{SIAM/ASA Journal on Uncertainty Quantification}, 2\penalty0 (1):\penalty0 245--251, 2014.

\bibitem[Owen and Prieur(2017)]{owen2017shapley}
Art~B Owen and Cl{\'e}mentine Prieur.
\newblock On shapley value for measuring importance of dependent inputs.
\newblock \emph{SIAM/ASA Journal on Uncertainty Quantification}, 5\penalty0 (1):\penalty0 986--1002, 2017.

\bibitem[Pearl(2009)]{pearl2009causality}
Judea Pearl.
\newblock \emph{Causality}.
\newblock Cambridge university press, 2009.

\bibitem[Pedregosa et~al.(2011)Pedregosa, Varoquaux, Gramfort, Michel, Thirion, Grisel, Blondel, Prettenhofer, Weiss, Dubourg, Vanderplas, Passos, Cournapeau, Brucher, Perrot, and Duchesnay]{scikit-learn}
F.~Pedregosa, G.~Varoquaux, A.~Gramfort, V.~Michel, B.~Thirion, O.~Grisel, M.~Blondel, P.~Prettenhofer, R.~Weiss, V.~Dubourg, J.~Vanderplas, A.~Passos, D.~Cournapeau, M.~Brucher, M.~Perrot, and E.~Duchesnay.
\newblock Scikit-learn: Machine learning in {P}ython.
\newblock \emph{Journal of Machine Learning Research}, 12:\penalty0 2825--2830, 2011.

\bibitem[Poelwijk et~al.(2019)Poelwijk, Socolich, and Ranganathan]{poelwijk_learning_2019}
Frank~J. Poelwijk, Michael Socolich, and Rama Ranganathan.
\newblock Learning the pattern of epistasis linking genotype and phenotype in a protein.
\newblock \emph{Nature Communications}, 10\penalty0 (1):\penalty0 4213, September 2019.
\newblock ISSN 2041-1723.
\newblock \doi{10.1038/s41467-019-12130-8}.
\newblock Number: 1 Publisher: Nature Publishing Group.

\bibitem[Rahaman et~al.(2019)Rahaman, Baratin, Arpit, Draxler, Lin, Hamprecht, Bengio, and Courville]{rahaman2019spectral}
Nasim Rahaman, Aristide Baratin, Devansh Arpit, Felix Draxler, Min Lin, Fred Hamprecht, Yoshua Bengio, and Aaron Courville.
\newblock On the spectral bias of neural networks.
\newblock In \emph{International Conference on Machine Learning}, pages 5301--5310. PMLR, 2019.

\bibitem[Rasmussen(2004)]{rasmussen2004gaussian}
Carl~Edward Rasmussen.
\newblock Gaussian processes in machine learning.
\newblock In \emph{Summer school on machine learning}, pages 63--71. Springer, 2004.

\bibitem[Ronen et~al.(2019)Ronen, Jacobs, Kasten, and Kritchman]{ronen2019convergence}
Basri Ronen, David Jacobs, Yoni Kasten, and Shira Kritchman.
\newblock The convergence rate of neural networks for learned functions of different frequencies.
\newblock \emph{Advances in Neural Information Processing Systems}, 32, 2019.

\bibitem[Sarkisyan et~al.(2016)Sarkisyan, Bolotin, Meer, Usmanova, Mishin, Sharonov, Ivankov, Bozhanova, Baranov, Soylemez, Bogatyreva, Vlasov, Egorov, Logacheva, Kondrashov, Chudakov, Putintseva, Mamedov, Tawfik, Lukyanov, and Kondrashov]{sarkisyan_local_2016}
Karen~S. Sarkisyan, Dmitry~A. Bolotin, Margarita~V. Meer, Dinara~R. Usmanova, Alexander~S. Mishin, George~V. Sharonov, Dmitry~N. Ivankov, Nina~G. Bozhanova, Mikhail~S. Baranov, Onuralp Soylemez, Natalya~S. Bogatyreva, Peter~K. Vlasov, Evgeny~S. Egorov, Maria~D. Logacheva, Alexey~S. Kondrashov, Dmitry~M. Chudakov, Ekaterina~V. Putintseva, Ilgar~Z. Mamedov, Dan~S. Tawfik, Konstantin~A. Lukyanov, and Fyodor~A. Kondrashov.
\newblock Local fitness landscape of the green fluorescent protein.
\newblock \emph{Nature}, 533\penalty0 (7603):\penalty0 397--401, May 2016.
\newblock ISSN 1476-4687.
\newblock \doi{10.1038/nature17995}.
\newblock Number: 7603 Publisher: Nature Publishing Group.

\bibitem[Scheibler et~al.(2015)Scheibler, Haghighatshoar, and Vetterli]{scheibler2015fast}
Robin Scheibler, Saeid Haghighatshoar, and Martin Vetterli.
\newblock A fast hadamard transform for signals with sublinear sparsity in the transform domain.
\newblock \emph{IEEE Transactions on Information Theory}, 61\penalty0 (4):\penalty0 2115--2132, 2015.

\bibitem[Shapley(1952)]{P-295}
Lloyd~S. Shapley.
\newblock \emph{A Value for N-Person Games}.
\newblock RAND Corporation, Santa Monica, CA, 1952.
\newblock \doi{10.7249/P0295}.

\bibitem[Shrikumar et~al.(2017)Shrikumar, Greenside, and Kundaje]{shrikumar2017learning}
Avanti Shrikumar, Peyton Greenside, and Anshul Kundaje.
\newblock Learning important features through propagating activation differences.
\newblock In \emph{International conference on machine learning}, pages 3145--3153. PMLR, 2017.

\bibitem[{\v{S}}trumbelj et~al.(2009){\v{S}}trumbelj, Kononenko, and {\v{S}}ikonja]{vstrumbelj2009explaining}
Erik {\v{S}}trumbelj, Igor Kononenko, and M~Robnik {\v{S}}ikonja.
\newblock Explaining instance classifications with interactions of subsets of feature values.
\newblock \emph{Data \& Knowledge Engineering}, 68\penalty0 (10):\penalty0 886--904, 2009.

\bibitem[Sundararajan and Najmi(2020)]{sundararajan2020many}
Mukund Sundararajan and Amir Najmi.
\newblock The many shapley values for model explanation.
\newblock In \emph{International conference on machine learning}, pages 9269--9278. PMLR, 2020.

\bibitem[Valle-Perez et~al.(2018)Valle-Perez, Camargo, and Louis]{valle2018deep}
Guillermo Valle-Perez, Chico~Q Camargo, and Ard~A Louis.
\newblock Deep learning generalizes because the parameter-function map is biased towards simple functions.
\newblock \emph{arXiv preprint arXiv:1805.08522}, 2018.

\bibitem[Van~den Broeck et~al.(2021)Van~den Broeck, Lykov, Schleich, and Suciu]{van2021tractability}
Guy Van~den Broeck, Anton Lykov, Maximilian Schleich, and Dan Suciu.
\newblock On the tractability of shap explanations.
\newblock In \emph{Proceedings of the 35th Conference on Artificial Intelligence (AAAI)}, 2021.

\bibitem[Voigt and Von~dem Bussche(2017)]{voigt2017eu}
Paul Voigt and Axel Von~dem Bussche.
\newblock The eu general data protection regulation (gdpr).
\newblock \emph{A Practical Guide, 1st Ed., Cham: Springer International Publishing}, 10\penalty0 (3152676):\penalty0 10--5555, 2017.

\bibitem[Williams(1996)]{williams1996computing}
Christopher Williams.
\newblock Computing with infinite networks.
\newblock \emph{Advances in neural information processing systems}, 9, 1996.

\bibitem[Wu et~al.(2016)Wu, Dai, Olson, Lloyd-Smith, and Sun]{wu_adaptation_2016}
Nicholas~C Wu, Lei Dai, C~Anders Olson, James~O Lloyd-Smith, and Ren Sun.
\newblock Adaptation in protein fitness landscapes is facilitated by indirect paths.
\newblock \emph{eLife}, 5:\penalty0 e16965, July 2016.
\newblock ISSN 2050-084X.
\newblock \doi{10.7554/eLife.16965}.
\newblock Publisher: eLife Sciences Publications, Ltd.

\bibitem[Yang and Salman(2019)]{yang2019fine}
Greg Yang and Hadi Salman.
\newblock A fine-grained spectral perspective on neural networks.
\newblock \emph{arXiv preprint arXiv:1907.10599}, 2019.

\bibitem[Yang(2021)]{yang2021fast}
Jilei Yang.
\newblock Fast treeshap: Accelerating shap value computation for trees.
\newblock \emph{arXiv preprint arXiv:2109.09847}, 2021.

\bibitem[Zern et~al.(2023)Zern, Broelemann, and Kasneci]{zern2023interventional}
Artjom Zern, Klaus Broelemann, and Gjergji Kasneci.
\newblock Interventional shap values and interaction values for piecewise linear regression trees.
\newblock In \emph{Proceedings of the AAAI Conference on Artificial Intelligence}, volume~37, pages 11164--11173, 2023.

\end{thebibliography}

\medskip
\clearpage
\appendix
\section{Relevant work}
\subsection{Neural network theory and simplicity biases}
\label{subsec:simplicity_bias_review}
With regards to simplicity biases in fully connected neural networks, a substantial amount of research has been dedicated to analyzing neural networks in function space. This line of research is dedicated to firstly showing that ``infinite-width'', randomly initialized (with Gaussian distribution) neural networks are distributed as Gaussian Processes (GPs) and secondly computing the kernel associated to the GP \citep{neal1996priors, williams1996computing, cho2009kernel, hazan2015steps,lee2017deep,ancona2018towards, daniely2017sgd}. The kernel associated with the GP is commonly known as the ``conjugate kernel'' \citep{daniely2017sgd} or the ``NN-GP kernel''\citep{lee2017deep}. Other works show that in infinite-width neural networks weights after training via SGD do not end up too far from the original \citep{chizat2019lazy, jacot2018neural, du2018gradient, allen2019convergence, allen2019convergencernn}, referred to as ``lazy training'' by \citet{chizat2019lazy}. This allowed \citet{jacot2018neural} to prove that the evolution of an infinite-width neural network during training can be described by a kernel called the ``Neural Tangent Kernel''. \citet{lee2019wide} showed, through extensive experiments that the same behavior holds even for the more realistic case of  neural nets of finite width. Empirically speaking, it was shown by \citet{rahaman2019spectral} that a neural net with one input tends to learn sinusoids of lower frequencies in earlier epochs than those with higher frequencies. By analyzing the spectrum of the NTK's Gram matrix, \citet{ronen2019convergence, basri2020frequency} were able to formally prove this empirical finding. \citet{yang2019fine, fan2020spectra} analyze the spectra of the NTK gram matrix for higher dimensional inputs.  Specifically,  \citet{yang2019fine, valle2018deep} provide simplicity bias results for the case where the inputs to the neural net are Boolean (zero-one) vectors. 
\subsection{Sparse and low-degree Fourier transform algorithms}
\label{subsec:sparse_fft_review}
We now discuss algorithms that efficiently approximate {\em general} black-box predictors by a Fourier sparse representation. Let $h:\{0, 1\}^n \rightarrow \R$ be a any  function. We assume we have query access to $h$. That is, we can arbitrarily pick $x \in \{0, 1\}^n$ and query $h$ for its value $h(x)$. Without any further assumptions, computing the Fourier transform requires us to query {\em exponentially}, to be precise $2^n$, many queries: one for every $x \in \{0,1\}^n$. Furthermore, classical Fast Fourier Transform (FTT) algorithms are known to take at least $\Omega(2^n \log(2^n))$ time. 

Under the additional assumption that $h$ is $k$-sparse, works such as \cite{cheraghchi2017nearly, amrollahi2019efficiently, scheibler2015fast, kushilevitz1993learning, li2015active, li2015spright} provide algorithms that obtain the Fourier transform more efficiently. In particular, \cite{amrollahi2019efficiently} provide algorithms with query complexity $O(nk)$ and time complexity $O(n k \log k)$ time. Assuming further that the function is of degree $d=o(n)$, the query complexity reduces to $O(k d \log n)$, with run time still polynomial in $n,k,d$. Crucially, even if the function $h$ is not $k$-sparse, Algorithm~\textsc{RobustSWHT} of \cite{amrollahi2019efficiently} yields the best $k$-sparse approximation in the $\ell_2- \ell_2$ sense. More precisely, let us denote by $h_k: \{0, 1\}^n \rightarrow$ the function that is formed by only keeping the top $k$ non-zero Fourier coefficients of $h$ and setting the rest to zero.  Then the algorithm returns a $O(k)$-sparse function $g$ such that:
\begin{align*}
&\sum\limits_{f \in \{0, 1\}^n} (\widehat{g}(f)-\widehat{h}(f))^2 \leq 
C(1+\epsilon) \min\limits_{\text{all } k\text{-sparse } g}  \sum\limits_{f \in \{0, 1\}^n} (\widehat{g}(f)-\widehat{h_k}(f))^2,
\end{align*}
where $C$ is a universal constant. By Parseval's identity, the same holds if the summations were over the time (input) domain instead of the frequency domain. 

\subsection{Shapley values in the context of Machine learning}
\label{subsec:shapley_literature_review}
In the context of ML, many works have derived a different notion of Shapley value depending on what they mean by data distribution, deleted features, etc.  We refer the reader to the survey by \citet{sundararajan2020many, pmlr-v108-janzing20a} for a comprehensive overview. In this work we focus on the notion of SHAP introduced by \citet{lundberg2017unified,lundberg2020local} also known as ``Interventional'' \citep{pmlr-v108-janzing20a,  van2021tractability} or ``Baseline'' SHAP  \citep{sundararajan2020many} where the missing features are integrated out from the marginal distribution as opposed to the conditional distribution (see Section~\ref{sec:background}). As pointed out by \citet{ pmlr-v108-janzing20a} there are two main ways to define the SHAP value ``interventional'' and ``observational'' SHAP. These are referred to by \citet{sundararajan2020many} as ``baseline'' and  ``conditional'' SHAP respectively. 

As pointed out by \citet{ pmlr-v108-janzing20a} the difference between these definitions can be better viewed with the lens of causality \citep{pearl2009causality}. Roughly speaking ``observational'' SHAP tells us about how influential a feature is to the prediction of the predictor if it goes from the state of being unobserved to observed.  ``Interventional'' SHAP is \emph{causal} and tells us how influential a feature is if we were to reach in (through a process called an intervention) and change that feature in order to change the prediction. Put into the context of credit scores and loan approvals, ``observational'' SHAP will provide us with important features which are ``observed'' by the predictor and hence are influential in predicting if a particular loan request will be rejected or approved. Interventional SHAP would tell us which feature we could change or ``intervene'' in order to change the outcome of the loan request. 

The original (ML) SHAP paper \citep{lundberg2017unified} proposes ``observational'' (conditional) SHAP as the correct notion of SHAP. \citet{van2021tractability, arenas2021tractability} provide intractability results for observational SHAP in a variety of simple distributional assumptions on the data and simple predictors $f$.
This line of research on observational SHAP has been extended in subsequent work, which provides computational theoretical results on the hardness of computing observational SHAP values for various classes of predictors and data distribution assumptions. 
\citep{huang2024updates, marzouk2024markovian, marzouk2025many}.
The definition of SHAP in the original work of \citep{lundberg2017unified} has also lead to many attempts to \emph{approximate} observational SHAP values \citep{lundberg2020local, covert2020improving, aas2021explaining, kwon2021efficient, sundararajan2020many}. It is interesting to note that the version of ``Kernel''-SHAP in  \citet{lundberg2017unified} is also an approximation for observational SHAP values that ends up coinciding precisely with interventional SHAP values, which explains a lot of the confusion in the community. \citet{ pmlr-v108-janzing20a} boldly claims that researchers should stop the pursuit of approximations to ``observation'' SHAP values as it lacks certain properties, for example, sensitivity i.e. the SHAP value of a feature can be non-zero while the predictor $f$ has no dependence on that feature. This phenomenon happens because when features are correlated, the presence of a feature can provide information about other features that the predictor does depend on. This does \emph{not} happen in interventional SHAP. Finally, \citet{chen2020true} argues that both SHAP definitions are worthy of pursuit. They emphasize that the interventional framework provides explanations that are more ``true to the model'', and the observational approach's explanations are more ``true to the data''. 

\section{Proofs}
\label{app:proofs}
Before we start with the proofs we review the Fourier analysis and synthesis equations. As we mentioned in the Background Section~\ref{sec:background}, the Fourier representation of the \emph{pseudo-boolean} function $h:\{0,1\}^n \rightarrow \R$ is the unique expansion of $h$ as follows: 
\[
h(x) = \frac{1}{\sqrt{2^n}}\sum\limits_{f \in \{0,1\}^n} \widehat{h}(f) (-1)^{\langle f,x \rangle}
\]This is the so called Fourier ``synthesis'' equation. 

The Fourier coefficients $\widehat{h}(f)$ are computed by the Fourier ``analysis'' equation:
\begin{equation}
\label{eqn:fourier_analysis}
\widehat{h}(f) =\frac{1}{\sqrt{2^n}}\sum\limits_{x \in \{0,1\}^n} h(x) (-1)^{\langle f,x \rangle}
\end{equation}

\subsection{Proof of propositions}
\label{app:proofs:propositions}
We can now prove Proposition~\ref{prop:decomposed_function}:
\propDecomposed*
\begin{proof}
Let $g:\{0,1\}^n  \rightarrow \mathbb{R}$ be a function dependent on exactly $d$ variables $x_{i_1}, \dots, x_{i_d}$, where $i_1, \dots, i_d \in [n]$ are distinct indices. We show that for any frequency $f \in \{0,1\}^n$, if $f_j =1$ for some $j \notin S \triangleq \{i_1, \dots, i_d\}$, then, $\widehat{g}(f)=0$. From the Fourier analysis Equation~\ref{eqn:fourier_analysis} we have:
\begin{align*}
\widehat{g}(f) 
&=\frac{1}{\sqrt{2^n}}\sum\limits_{x \in \{0,1\}^n} g(x) (-1)^{\langle f,x \rangle} = \sum\limits_{x_i: i \in S} \sum\limits_{x_j: j \in [n] \setminus S} g(x) (-1)^{\langle f_S,x_S \rangle} (-1)^{\langle f_{[n] \setminus S},x_{[n] \setminus S} \rangle} 
\\&
\labelrel={app:eq:sum1}  \sum\limits_{x_i: i \in S} g(x) (-1)^{\langle f_S,x_S \rangle} \sum\limits_{x_j: j \in [n] \setminus S} (-1)^{\langle f_{[n] \setminus S},x_{[n] \setminus S} \rangle} \labelrel={app:eq:sum2} \sum\limits_{x_i: i \in S} g(x) (-1)^{\langle f_S,x_S \rangle} \cdot 0 = 0
\end{align*}

Where Equation~\ref{app:eq:sum1} holds because $g$ is only dependent on the variables in $S$ and Equation~\ref{app:eq:sum2} holds by checking the inner sum has an equal number of $1$ and $-1$ added together.

The proof of the proposition follows by the linearity of the Fourier transform.
\end{proof}
Moving on to Proposition~\ref{prop:degree_imlies_sparsity}:
\propDegree*
\begin{proof}
We simply note that the number of frequencies $f \in \{0,1\}^n$ of degree at most $d$ is equal to $\sum\limits_{i=0}^d \binom{n}{i}$. This sum is $O(n^d)$ for $d$ constant w.r.t $n$. 
\end{proof}
\subsection{Proof of Lemma \ref{lemma:shapf_fourier_basis}}
\label{app:proofs:lemma1}
\begin{proof}
We start from Equation~\ref{eq:shapf1}:
\begin{align*}
\phi_i^{\Psi_f} &= 
\sum\limits_{S \subseteq [n] \setminus \{i\}}  \frac{|S|! (n-|S|-1)!}{n!} \cdot \frac{1}{|\mathcal{D}|} \sum\limits_{(x,y) \in \mathcal{D}} 
\left((-1)^{\langle f, x^*_{S \cup\{i\}} \oplus x_{[n] \setminus S \cup \{i\}} \rangle} - (-1)^{\langle f, x^*_S \oplus x_{[n] \setminus S}\rangle}\right) \nonumber \\
&= 
\frac{1}{|\mathcal{D}|} \sum\limits_{(x,y) \in \mathcal{D}} \sum\limits_{S \subseteq [n] \setminus \{i\}}  \frac{|S|! (n-|S|-1)!}{n!} 
(-1)^{\langle f_{-i}, x^*_S \oplus x_{[n] \setminus S \cup \{i\}}\rangle} \left((-1)^{f_ix^*_i} - (-1)^{f_ix_i}\right) 
\end{align*}

By checking all 8 possible combinations of $x_i, x_i^*, f_i \in \{0,1\}$, one can see that $(-1)^{f_ix^*_i} - (-1)^{f_ix_i} = 2f_i(x_i-x_i^*)$. This is simply because the two exponents only differ when $f_i=1$ and $x_i\neq x_i^*$.

To determine $(-1)^{\langle f_{-i}, x^*_S \oplus x_{[n] \setminus S \cup \{i\}}\rangle}$, we partition $[n] \setminus \{i\}$ into two subsets $A \triangleq \{j \in [n] | x_j \neq x^*_j,  j \neq i, f_j=1\}$ and $B \triangleq [n] \setminus A \cup \{i\}$. Doing this, we can factor out  $(-1)^{\langle f_{-i}, x_{-i}\rangle}$ and determine the rest of the sign based on the number of indices in $S$ where $x$ and $x^*$ disagree and $f_i=1$. This is equal to $|A \cap S|$:
\begin{equation*}
\phi_i^{\Psi_f} =  
\frac{2f_i}{|\mathcal{D}|} \sum\limits_{(x,y) \in \mathcal{D}}
(x_i - x^*_i)(-1)^{\langle f_{-i}, x_{-i}\rangle} 
\sum\limits_{S \subseteq [n] \setminus \{i\}}  \frac{|S|! (n-|S|-1)!}{n!} 
(-1)^{|A \cap S|}
\end{equation*}
$A$ and $B$ partition $[n] \setminus \{i\}$, therefore we split the inner sum as follows:
\begin{equation*}
\phi_i^{\Psi_f} =  \frac{2f_i}{|\mathcal{D}|} \sum\limits_{(x,y) \in \mathcal{D}}
(x_i - x^*_i)(-1)^{\langle f_{-i}, x_{-i}\rangle}
\sum\limits_{\Tilde{B} \subseteq B} \sum\limits_{\Tilde{A} \subseteq A}  \frac{(|\Tilde{A}| + |\Tilde{B}|)! (n-(|\Tilde{A}| + |\Tilde{B}|)-1)!}{n!} 
(-1)^{|\Tilde{A}|}
\nonumber  \\
\end{equation*}
Since the inner expression only depends on the cardinalities of $\Tilde{A}$ and $\Tilde{B}$ we can recast the inner sum to be over numbers instead of subsets by counting the number of times each cardinality appears in the summation:
\begin{align}
\label{app:eq:shapf_simp}
\phi_i^{\Psi_f} &= \frac{2f_i}{|\mathcal{D}|} \sum\limits_{(x,y) \in \mathcal{D}}
(x_i - x^*_i)(-1)^{\langle f_{-i}, x_{-i}\rangle} 
\sum_{b = 0}^{n - |A| - 1} \sum_{a = 0}^{|A|} \binom{n - |A| - 1}{b} \binom{|A|}{a}  \frac{(a + b)! (n-a-b-1)!}{n!} 
(-1)^{a} \nonumber \\
 &= \frac{2f_i}{|\mathcal{D}|} \sum\limits_{(x,y) \in \mathcal{D}} 
 (x_i - x^*_i)(-1)^{\langle f_{-i}, x_{-i}\rangle} 
\sum_{a = 0}^{|A|} (-1)^{a}
\sum_{b = 0}^{n - |A| - 1} \frac{\binom{|A|}{a}\binom{n - |A| - 1}{b}}{n \binom{n  - 1}{a + b}}  
\end{align}
Now we find a closed-form expression for the innermost summation in the above Equation, which is a summation over $b$ where $a$ is fixed:
\begin{align}
\sum_{b = 0}^{n - |A| - 1} \frac{\binom{|A|}{a}\binom{n - |A| - 1}{b}}{n\binom{n  - 1}{a + b}} & \labelrel={app:eq:comb1} 
\sum_{b = 0}^{n - |A| - 1} \frac{\binom{|A|}{a}\binom{n - |A| - 1}{b}}{
n \frac{\binom{n-1}{|A|}}{\binom{a+b}{a} \binom{n-a-b-1}{|A|-a}}
\binom{|A|}{a}\binom{n - |A| - 1}{b}} \nonumber \\ 
 &=\frac{1}{n \binom{n-1}{|A|}}
\sum_{b = 0}^{n - |A| - 1} \binom{a+b}{a} \binom{n-a-b-1}{|A|-a} \nonumber \\ 
& \labelrel={app:eq:comb2} \frac{1}{n \binom{n-1}{|A|}}
\binom{n}{|A|+1} \nonumber \\
&= \frac{1}{|A|+1} 
\label{app:eq:summation}
\end{align}
In Equation~\ref{app:eq:comb1}, we use the following identity: $\binom{n  - 1}{a + b}\binom{a+b}{a} \binom{n-a-b-1}{|A|-a} = 
\binom{n-1}{|A|} \binom{|A|}{a}\binom{n - |A| - 1}{b}$. This can be checked algebraically by simply writing down each binomial term as factorials and doing the cancellations:

\begin{align*}
& \binom{n-1}{a+b} \binom{a+b}{a} \binom{n-a-b-1}{|A|-a} \\
& = \frac{(n-1)!}{(a+b)!(n-a-b-1)!} \cdot \frac{(a+b)!}{a!b!} \cdot \frac{(n-a-b-1)!}{(|A|-a)!(n-a-b-1-(|A|-a))!} \\
& = \frac{(n-1)!}{a!b!(|A|-a)!(n-|A|-b-1)!}. \\
& = \frac{(n-1)!}{|A|!(n-|A|-1)!} \cdot \frac{|A|!}{a!(|A|-a)!} \cdot \frac{(n-|A|-1)!}{b!(n-|A|-1-b)!} \\
& = \binom{n-1}{|A|} \binom{|A|}{a} \binom{n-|A|-1}{b}
\end{align*}

In Equation~\ref{app:eq:comb2}, we use $\sum_{b = 0}^{n - |A| - 1} \binom{a+b}{a} \binom{n-a-b-1}{|A|-a}= \binom{n}{|A|+1}$ which holds because of the following double-counting argument. 
The term $\binom{n}{|A|+1}$ counts the number of ways to choose a subset of size $|A|+1$ from a set of $n$ elements. 
Imagine elements are numbered from 1 to n, and $\pi_1 < \pi_2 < ... < \pi_{|A|+1}\in [n]$ represent $|A|+1$ chosen elements. Let's condition on $\pi_{a+1}=(a+b+1)$ where possible values for $b$ can only be $0 \leq b \leq n-|A|-1$. This is due to the fact that $\pi_{a+1}<a+1$ implies $\pi_1, ..., \pi_a$ are chosen from less than $a$ elements, and similarly $\pi_{a+1}>n-|A|+a$ implies $\pi_{a+2}, ... ,\pi_{|A|+1}$ ($|A|-a$ chosen elements) are chosen from less than $|A|-a$ elements, which are both impossible.
Given the condition, $a$ elements numbered lower than $(a+b+1)$ and $|A|-a$ elements numbered larger than $(a+b+1)$ are also chosen. This is possible in $\binom{a+b}{a} \binom{n-a-b-1}{|A|-a}$ ways. Therefore, keeping in mind that $a$ is fixed, $\sum_{b = 0}^{n - |A| - 1} \binom{a+b}{a} \binom{n-a-b-1}{|A|-a}$ should give the number of ways to choose a subset of size $|A|+1$ from a set of $n$ elements, through the new perspective.

Based on Equation~\ref{app:eq:summation}, we see that the innermost summation in Equation~\ref{app:eq:shapf_simp} is only dependent on $|A|$. Thus, we rewrite Equation~\ref{app:eq:shapf_simp} as follows:
\begin{align*}
\phi_i^{\Psi_f}  &= \frac{2f_i}{|\mathcal{D}|} \sum\limits_{(x,y) \in \mathcal{D}} (x_i - x^*_i)(-1)^{\langle f_{-i}, x_{-i}\rangle} 
\frac{(|A|+1) \mod 2}{|A|+1}
\end{align*}

By absorbing the sign of $(x_i-x_i^*)$ into the $(-1)^{{\langle f_{-i}, x_{-i}\rangle}}$ term we arrive at Equation~\ref{eq:shapf_fourier_basis}:
\begin{equation*}
\phi_i^{\Psi_f}  = -\frac{2f_i}{|\mathcal{D}|} \sum\limits_{(x,y) \in \mathcal{D}}  \mathmybb{1}_{x_i \neq x^*_i}(-1)^{\langle f, x\rangle}  \frac{(|A|+1) \mod 2}{|A|+1} 
\end{equation*}
\end{proof}
\subsection{Proof of Theorem \ref{thm:main}}
\label{app:proofs:thm2}
\begin{proof}
Proof of Equation~\ref{eq:shapf_final} simply follows from the fact that SHAP values are linear w.r.t. the explained function. Regarding the computational complexity we restate Equation~\ref{eq:shapf_final}
\begin{equation*}
\phi_i^{h}  = -\frac{2}{|\mathcal{D}|} \sum\limits_{f \in \text{supp}(h)} \widehat{h}(f) \cdot f_i \sum\limits_{(x,y) \in \mathcal{D}}  \mathmybb{1}_{x_i \neq x^*_i}(-1)^{\langle f, x\rangle}  \frac{(|A|+1) \mod 2}{|A|+1} 
\end{equation*}
where $A \triangleq \{j \in [n] | x_j \neq x^*_j,  j \neq i, f_j=1\}$. 

We first do a pre-processing step for amortizing the cost of computing $|A|$: we compute $\tilde{A} \triangleq \{j \in [n] | x_j \neq x^*_j,  f_j=1\}$ which takes $\Theta(|\mathcal{D}|n)$ flops. 

We assume we are computing the whole vector $\Phi^h = (\Phi^h_1, \dots, \Phi^h_n)$, that is we are compute SHAP values for all $i \in [n]$ at the same time. Going back to the inner summation above, computing $A$ (and $|A|$) for different values of $i \in [n]$ is $\Theta(n)$ if we utilize the pre-computed $\tilde{A}$.  The inner product $\langle f, x \rangle$ is not dependent on $i$ and is $\Theta(n)$ flops.  Computing $\mathmybb{1}_{x_i \neq x^*_i}$ for different values of $i \in [n]$ is $\Theta(n)$. Therefore, the inner expression of the summand takes $\Theta(n)$ flops for a fixed data-point $x \in \mathcal{D}$ and $f \in \text{supp}(h)$. 

Computing the inner sum for any fixed frequency $f \in \text{supp}(h)$ is $\Theta(n |\mathcal{D}|)$, because we are summing over $|\mathcal{D}|$ vectors each of size $n$ (the vector which holds SHAP value for each $i \in [n]$). Moving on to the outer sum each evaluation of the inner sum is $\Theta(n |\mathcal{D}|)$ and it results in  a vector of size $n$ (one element for each SHAP value). The multiplication of $ \widehat{h}(f) \cdot f_i $ is $\Theta(n)$. Therefore the cost of the inner sum dominates i.e.  $\Theta(n |\mathcal{D}|)$. Since we are summing over the whole support the total number of flops is: $\Theta(n |\mathcal{D}| k )$ where $k=|\text{supp}(h)|$.
\end{proof}
\section{Datasets}
\label{app:sec:datasets}
We list all the datasets used in the Experiments Section~\ref{sec:experiments}.
\paragraph{Entacmaea quadricolor fluorescent protein. (Entacmaea)}
\cite{poelwijk_learning_2019} study the fluorescence brightness of all $2^{13}$ distinct variants of the Entacmaea quadricolor fluorescent protein, mutated at $13$ different sites. 

\paragraph{GPU kernel performance (SGEMM).}
\cite{nugteren_cltune_2015} measures the running time of a matrix product using a parameterizable SGEMM GPU kernel, configured with different parameter combinations. The input has $14$ categorical features. After one-hot encoding the dataset is $40$-dimensional.

\paragraph{Immunoglobulin-binding domain of protein G (GB1).}
\cite{wu_adaptation_2016} study the ``fitness'' of variants of protein GB1, that are mutated at four different sites. Fitness, in this work, is a quantitative measure of the stability and functionality of a protein variant. Given the $20$ possible amino acids at each site, they report the fitness for $20^4=160,000$ possible variants, which we represent with one-hot encoded 80-dimensional binary vectors. 

\paragraph{Green fluorescent protein from Aequorea victoria (avGFP).}
\cite{sarkisyan_local_2016} estimate the fluorescence brightness of random mutations over the green fluorescent protein sequence of Aequorea victoria (avGFP) at 236 amino acid sites. We transform the amino acid features into binary features indicating the absence or presence of a mutation at each amino acid site. This converts the original $54,024$ distinct amino acid sequences of length 236 into $49,089$ 236-dimensional binary data points.

\section{FourierShap Implementation}
\label{app:implementation}
We implemented four versions of our algorithm, i.e., Equation~\ref{eq:shapf_final}, using Google JAX library~\citep{jax2018github}. JAX provides a flexible framework for developing high-performace functions for vectorized computations. JAX enables automatic performance optimisation of algebraic computation as well as just-in-time (JIT) compilation of the vectorized functions for faster iterations at runtime.

The full code base is provided as supplementary material to the paper. We refer the reader there for more details. Here we give a high level overview of the four versions we implemented and experimented with:
\begin{itemize}
    \item \textbf{Base}: In this version, frequencies are represented as $n$-dimensonal binary vectors. A simple implenetation is provided for computing the SHAP values given a single frequency, a single background instance, and a single query instance. This is next extended to multiple frequencies, multiple background instances, and multiple query instances using multiple JAX \verb|vmap|s.
    \item \textbf{Precompute}: This version is a modified version of the Base version.  We take a closer look at Equation~\ref{eq:shapf_final} and pre-compute terms dependent only on frequencies $f$ and background dataset points $x$ (not including terms dependent on the query point $x^*$). These are then loaded from memory at run-time. This version was faster than Base in all of our experiments, but inherently requires more memory; in our case, more GPU memory.
    \item \textbf{Sparse}: Here we utilize the sparsity of frequencies $f \in \{0,1\}^n$ i.e. the fact that frequencies have mostly zero entries in practice. In this version, for each frequency $f$, we focus on positions $i$ where $f_i=1$. Inspecting Equation~\ref{eq:shapf_final}, these are the only positions where $f$ can affect the final SHAP value vector.
    \item \textbf{Positional}: This version brings again builds on the previous and computes SHAP values separately for each coordinate $i \in [n]$. This adds one extra precomputation step to map frequencies to coordinates they affect, but enables mathematical simplifications as well as potential for extra vectorization. We observed that although this extra precomputation step could become time-consuming for large set of frequencies, it can result in a significantly faster SHAP value computation at run-time in the case of low-degree Fourier spectrums, compared to the Sparse version.
\end{itemize}
\section{Experiment Details}
The code for running the experiments and the implementations of all modules are open-sourced and is publicly available at \url{https://github.com/andisheh94/fouriershap}. We run all experiments on a machine with one NVIDIA GeForce RTX 4090 GPU, on servers with Intel(R) Xeon(R) CPU E3-1284L v4 @ 2.90GHz, restricting the memory/RAM to 20 GB, which was managed with Slurm. 

\subsection{Black-box}
\label{app:experiments:blackbox}
\begin{figure*}[!htb]
    \centering
     \begin{subfigure}[b]{0.48\linewidth}
         \centering
         \includegraphics[width=\linewidth]{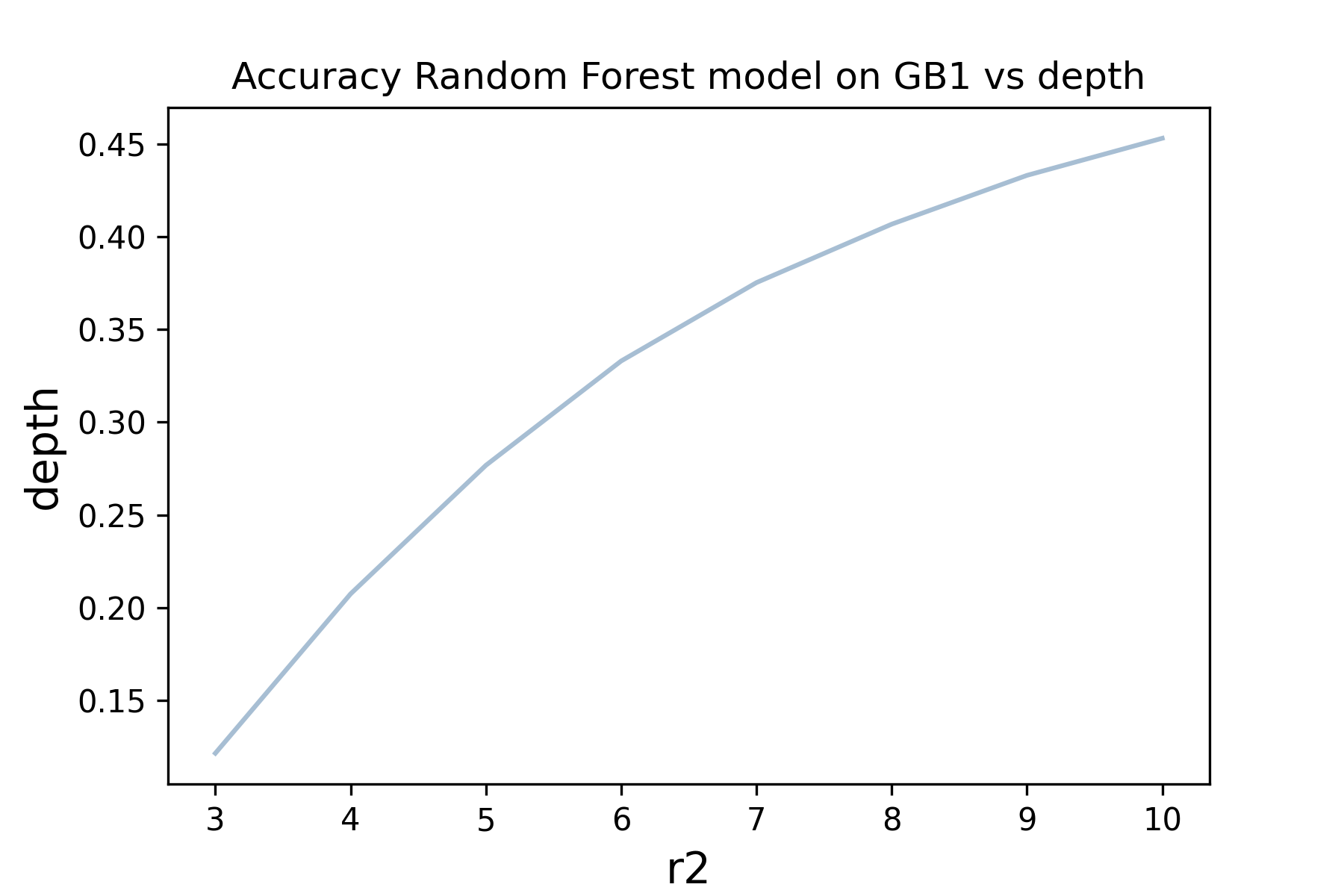}
         \label{gb1_model_accuracy}
     \end{subfigure}
     \begin{subfigure}[b]{0.48\linewidth}
         \centering
         \includegraphics[width=\linewidth]{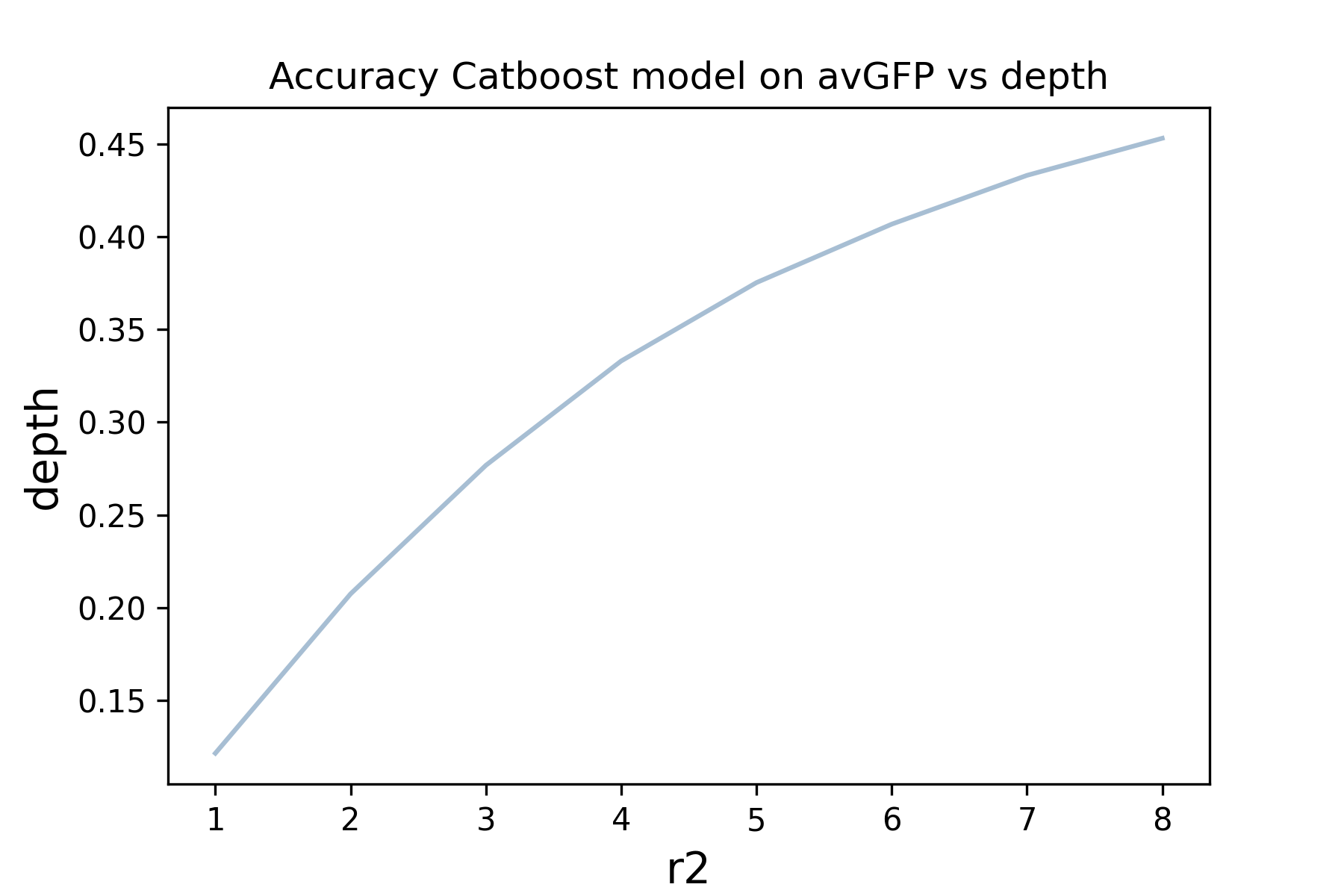}
         \label{avGFP_model_accuracy}
     \end{subfigure}
     \vspace*{-4mm}
    \caption{Model accuracy of tree models evaluated on the test set  for different depths}
\label{fig:model_accuracy}
\end{figure*}

For the Entacmaea and SGEMM datasets, we train fully connected neural networks  with 3 hidden layers containing 300 neurons each. The network is trained using the means-squared loss and ADAM optimizer with a learning rate of 0.01.  
For GB1 we train ensembles of trees models of varying depths using the random forest algorithm using the sklearn library \citep{scikit-learn}. For avGFP we train again, ensembles of trees models with 10 trees of varying depths using the cat-boost algorithm/library\citep{dorogush2018catboost}. All other setting are set to the default in both cases. Model accuracy for different depths are plotted in Figure~\ref{fig:model_accuracy}.  

For each dataset, we use all possible combination of number of background samples and query instances (instances to be explained) from $\{10, 20, 30, 40\}$, resulting in a total of 16 runs per dataset. Error bars in Figure~\ref{fig:speedup} capture the variation in speedups. 

\subsubsection{FourierSHAP}
For the first step of the \textsc{FourierSHAP}, i.e., computing a sparse Fourier approximation of the black-box model, we use an implementation of a sparse Walsh-Hadamard Transform (sparse WHT) a.k.a Fourier transform algorithm \citep{amrollahi2019efficiently} for each of the four trained models. The implementation is done using Google JAX~\citep{jax2018github} library as part of this work, and is available in the project repository.

For the second step of \textsc{FourierSHAP}, i.e., utilizing the approximated Fourier spectrum to compute \textsc{SHAP} values as per Equation~\ref{eq:shapf_final}, we use our ``Precompute'' implementation. We discussed the details in the previous Section, Section~\ref{app:implementation}.

To showcase the flexibility of our method in controlling the trade-off between speed and accuracy in Figure~\ref{fig:speedup}, we prune the computed sparse Fourier spectrum in the first step, and remove frequencies with amplitudes smaller than a specific threshold from the spectrum. For the threshold, we use $10$ values between $0.0001$ and $0.05$ and specified a minimal descriptive subset of them as points in Figure ~\ref{fig:speedup}.



\subsubsection{KernelShap}
For \textsc{KernelShap}, we use the standard library provided by the writers of the paper \citep{lundberg2017unified} \footnote{\url{https://github.com/slundberg/shap}} with its default settings. As part of the default setting, the paired sampling trick is also enabled, which shown to be beneficial for faster convergence of \textsc{KernelSHAP}. \textsc{KernelShap} is written in C and is to our knowledge the fastest implementation of this algorithm. 

To ensure a fair comparison, we run models on GPU wherever possible, i.e., for Neural Networks on Entacmaea and SGEMM datasets.


To emphasize the practical significance of speedups of \textsc{FourierSHAP} over \textsc{KernelSHAP}, we report the absolute runtimes of \textsc{KernelSHAP} (in seconds) in our experiment on the SGEMM dataset in Table~\ref{tab:kernel_shap_runtime_sgemm}. The runtimes correspond to the setting used in the main text, where background sizes were kept relatively small. This decision was made intentionally; during experimentation, we observed that larger background sizes caused \textsc{LinregSHAP} and \textsc{KernelSHAP} to become prohibitively slow, making it impractical to include full-scale results across datasets and seeds. Despite this limitation, our method already achieved orders-of-magnitude speedups.

\begin{table}[h!]
\centering
\begin{tabular}{lcccc}
\toprule
\textbf{Query Size $\downarrow$ / Background Size $\rightarrow$} & \textbf{10} & \textbf{20} & \textbf{30} & \textbf{40} \\
\midrule
\textbf{10} & 2.0 s & 3.9 s & 5.6 s & 7.8 s \\
\textbf{20} & 2.9 s & 5.8 s & 8.7 s & 11.7 s \\
\textbf{30} & 3.9 s & 7.5 s & 11.0 s & 14.4 s \\
\textbf{40} & 4.9 s & 9.6 s & 14.0 s & 17.9 s \\
\bottomrule
\end{tabular}
\caption{\textsc{KernelSHAP} runtimes in seconds in our experiment on the SGEMM dataset across varying sizes of the query and background sets.}
\label{tab:kernel_shap_runtime_sgemm}
\end{table}

To further emphasize the practical gains of \textsc{FourierSHAP}, we conducted additional experiments with significantly larger query and background sets, closer to what might be used in real-world deployments. Table~\ref{tab:kernel_shap_runtime_large} reports the runtimes in this setting. Despite the background dataset still being modest in size (100–400), \textsc{KernelSHAP} runtimes quickly increase to minutes, while \textsc{FourierSHAP} continues to run in milliseconds. These results reinforce that our reported relative speedups (often 100×–10,000×) translate into meaningful real-world runtime improvements.

\begin{table}[h!]
\centering
\begin{subtable}{\textwidth}
\centering
\begin{tabular}{lcccc}
\toprule
\textbf{Query Size $\downarrow$ / Background Size $\rightarrow$} & \textbf{100} & \textbf{200} & \textbf{300} & \textbf{400} \\
\midrule
\textbf{100} & 35 s & 71 s & 104 s & 133 s \\
\textbf{200} & 43 s & 84 s & 129 s & 169 s \\
\textbf{300} & 52 s & 107 s & 161 s & 212 s \\
\textbf{400} & 69 s & 132 s & 197 s & 265 s \\
\bottomrule
\end{tabular}
\caption{\textsc{KernelSHAP} runtimes (in seconds)}
\end{subtable}

\begin{subtable}{\textwidth}
\centering
\begin{tabular}{lcccc}
\toprule
\textbf{Query Size $\downarrow$ / Background Size $\rightarrow$} & \textbf{100} & \textbf{200} & \textbf{300} & \textbf{400} \\
\midrule
\textbf{100} & 14 ms & 28 ms & 43 ms & 57 ms \\
\textbf{200} & 29 ms & 58 ms & 86 ms & 110 ms \\
\textbf{300} & 40 ms & 79 ms & 119 ms & 159 ms \\
\textbf{400} & 53 ms & 106 ms & 160 ms & 213 ms \\
\bottomrule
\end{tabular}
\caption{\textsc{FourierSHAP} runtimes (in milliseconds)}
\end{subtable}
\caption{\textsc{KernelSHAP} runtimes vs. \textsc{FourierSHAP} runtimes across significantly larger sizes of the query and background sets than the setting used in the experiment on the SGEMM dataset.}
\label{tab:kernel_shap_runtime_large}
\end{table}

\subsubsection{Fast-SHAP}
\label{app:fastshap}
We tried our best to capture the full potential of \textsc{FastSHAP} in predicting SHAP values in terms of speed and accuracy, to ensure a practical and grounded comparison to our method.  
Here are the details on training Fast-SHAP as a baseline for computing interventional SHAP values:
\begin{itemize}
    \item \textbf{Imputer:} "Imputer" is a FastSHAP module used in the computation of neural networks' loss, acting as the value function in SHAP formula, that generates model's prediction using a strategy for treating features excluded in the subset, given the predictor and a subset of features. To compute interventional SHAP values, we use \verb|MarginalImputer|, implemented in the original FastSHAP repo, which computes mean predictions when using the backgorund dataset's values for excluded features. Therefore, each trained FastSHAP model is specific to a fixed background dataset, as the Imputer used in its loss is. In our experiments, we use four different background datasets with multiple sizes per (real-world) dataset, which lead to training four FastSHAP models per setting.

    \item \textbf{Feature Subset Sampling:} We train FastSHAP models with $1$, $4$, and $16$ feature subset samples per input. Although we did not observe monotonic improvements by increasing the number of feature subset samples, we decided to use the models trained with $16$ samples per input to compare our method with, as it was mostly performing the best in terms of accuracy.

    \item \textbf{Paired Sampling:} We enable paired sampling in FastSHAP, which is a trick to pair each feature subset sample $s \in \{0,1\}^n$ with its complement $1-s$, that is shown to be beneficial in reducing the variance and improving the accuracy, in both KernelSHAP and FastSHAP.

    \item \textbf{Neural Network Architectures:} We use MLPs with three different sizes to train FastSHAP, and reported the results for each separately:

    \begin{itemize}
        \item Small (2-layer): $in \times 128 \times out$.
        \item Medium (4-layer): $in \times 128 \times 128 \times 128 \times out$.
        \item Large (6-layer): $in \times 128 \times 128 \times 128 \times 128 \times 128 \times out$.
    \end{itemize}
    We use ReLU as the activation function in all models.

    \item \textbf{Hyperparameters:} We use the training batch size of $64$ and FastSHAP's defaults for all other components and hyper-parameters. We train each model with early stopping and up to $200$ epochs.
\end{itemize}


Table \ref{tab:fastshap_training_time} shows the time required to train the \textsc{FastSHAP} models used in our black-box experiments. Both \textsc{FourierShap} (ours) and \textsc{FastSHAP} are amortized methods that have a heavier ``pre-computation'' step. For \textsc{FastShap} this pre-computation appears as training an MLP that directly predicts SHAP values and for us this appears as computing a Fourier transform. When comparing \textsc{FastShap}'s initial training time to ours (reported in Figure~\ref{fig:fourier_runtime}), we can see our method has a considerably lower pre-computation time.

\begin{table}[h!]
\centering
\begin{tabularx}{\textwidth}{c c c |>{\centering\arraybackslash}X>{\centering\arraybackslash}X>{\centering\arraybackslash}X>{\centering\arraybackslash}X|c}
\toprule
\multirow{2}{*}{\textbf{Dataset}} & \multirow{2}{*}{\textbf{\makecell{Black-box \\ model}}} & \multirow{2}{*}{\textbf{\makecell{FastSHAP \\ size}}} & \multicolumn{4}{c|}{\textbf{\makecell{Training time (m) \\ for background dataset size}}} & \multirow{2}{*}{\textbf{\makecell{Total training \\ time (m)}}} \\ 
\cmidrule(lr){4-7}
& & & \textbf{10} & \textbf{20} & \textbf{30} & \textbf{40} & \\ 
\midrule
\multirow{3}{*}{Entacmaea} & \multirow{3}{*}{MLP} & Small & 5 & 8 & 15 & 18 & 46 \\
& & Medium & 2 & 4 & 6 & 10 & 22 \\
& & Large & 3 & 6 & 11 & 11 & 31 \\
\midrule
\multirow{3}{*}{SGEMM} & \multirow{3}{*}{MLP} & Small & 47 & 130 & 249 & 173 & 599 \\
& & Medium & 55 & 52 & 162 & 128 & 397 \\
& & Large & 72 & 73 & 152 & 91 & 388 \\
\midrule
\multirow{3}{*}{GB1} & \multirow{3}{*}{\makecell{Random Forest \\ (depth=5)}} & Small & 114 & 148 & 42 & 173 & 477 \\
& & Medium & 83 & 207 & 55 & 130 & 475 \\
& & Large & 54 & 102 & 155 & 184 & 495 \\
\midrule
\multirow{3}{*}{avGFP} & \multirow{3}{*}{\makecell{Catboost \\ (depth=5)}} & Small & 24 & 28 & 36 & 39 & 127 \\
& & Medium & 35 & 58 & 41 & 62 & 196 \\
& & Large & 11 & 20 & 22 & 25 & 78 \\
\bottomrule
\end{tabularx}
\caption{Training times of \textsc{FastSHAP} models used in our black-box experiments (in minutes). Unlike \textsc{FourierSHAP} (ours), \textsc{FastSHAP} needs to be separately trained for each background dataset. Part of the difference in the training times are due to the variance in the number of training epochs before the early stopping occurs. Original datasets are also of varying sizes that lead to different number of training samples per epoch.}
\label{tab:fastshap_training_time}
\end{table}

\subsubsection{LinRegShap}
\textsc{LinRegShap}, is a variance-reduced version of \textsc{KernelSHAP} \citep{covert2020improving}. We again use the implementation of the original writers\footnote{\url{https://github.com/iancovert/shapley-regression}}. Their implementation  includes automatic detection of the convergence of stochastic sampling which is meant to speed up the algorithm by taking less samples from the black box. Furthermore, as per the default setting, we also allowed paired sampling.

To ensure a fair comparison, we run models on GPU wherever possible, i.e., for Neural Networks on Entacmaea and SGEMM datasets.

\subsubsection{DeepLift}
For DeepLift we as well use the library of \citet{lundberg2017unified} \footnote{\url{https://github.com/slundberg/shap}} with default settings. In this setting the neural network is passed to the algorithm on a GPU to make sure this algorithm is as fast as possible.

\subsection{Trees}
\label{app:experiments:whitebox}
We fit random forests of maximum depths ranging from 3 to 8 with $20$ estimators on $90\%$ of all four datasets used in the black-box setting.
We compare performance of four algorithms in computing SHAP values for these random forest models;  
the classic \textsc{TreeSHAP} as well as its GPU implementation~\footnote{\url{https://github.com/slundberg/shap}}, \textsc{FastTreeSHAP}~\footnote{\url{https://github.com/linkedin/FastTreeSHAP}}, and our \textsc{FourierSHAP}. 
We always use $100$ datapoints as the background data samples, and $100$ datapoints to explain and compute the SHAP values for. We report the speedup of each algorithm over classic \textsc{TreeSHAP} in Figure~\ref{fig:whitebox}, with error bars showing the standard deviation in speedup over five independent runs.

For the first step of the \textsc{FourierSHAP}, we derive the exact sparse Fourier representation from the ensemble of trees, accessing the tree structures and using Equation~\ref{eq:recursive_tree}. We also perform a pruning on the exact sparse Fourier representation derived from the ensemble of trees, and keep the frequencies with largest amplitudes that cover at least $99.95\%$ of the original Fourier spectrum's energy. We also make sure to only remove frequencies with apmlitudes smaller than $0.005$.

To run \textsc{FourierSHAP}, we use our ``Precompute'' implementation for Entacmaea, and ``Positional'' version for other datasets, given larger feature spaces and the low-degreeness of the spectrum as a result of bounded maximum depth. See Section~\ref{app:implementation} for implementation details. We compare the resulted SHAP values with \textsc{TreeSHAP} results and achieve $R^2$ of at least $0.99$, ensuring precise values computed by \textsc{FourierSHAP}.

\subsection{Using KernelShap to produce ground truth SHAP values}
\label{app:kernel_shap_is_ground_truth}
As mentioned before, in order to measure the accuracy of the SHAP values produced by our and other algorithms we
need ground truth SHAP values. We use \textsc{KernelShap} to this extent. \textsc{KernelShap}
is an approximation method for computing interventional SHAP values. This
approximation becomes more precise by sampling more coalition subsets. Therefore, to generate
ground truth values, for each dataset, we sample more and more and check for convergence in these
values.


\begin{figure*}[!t]
    \centering
    \includegraphics[width=\linewidth]{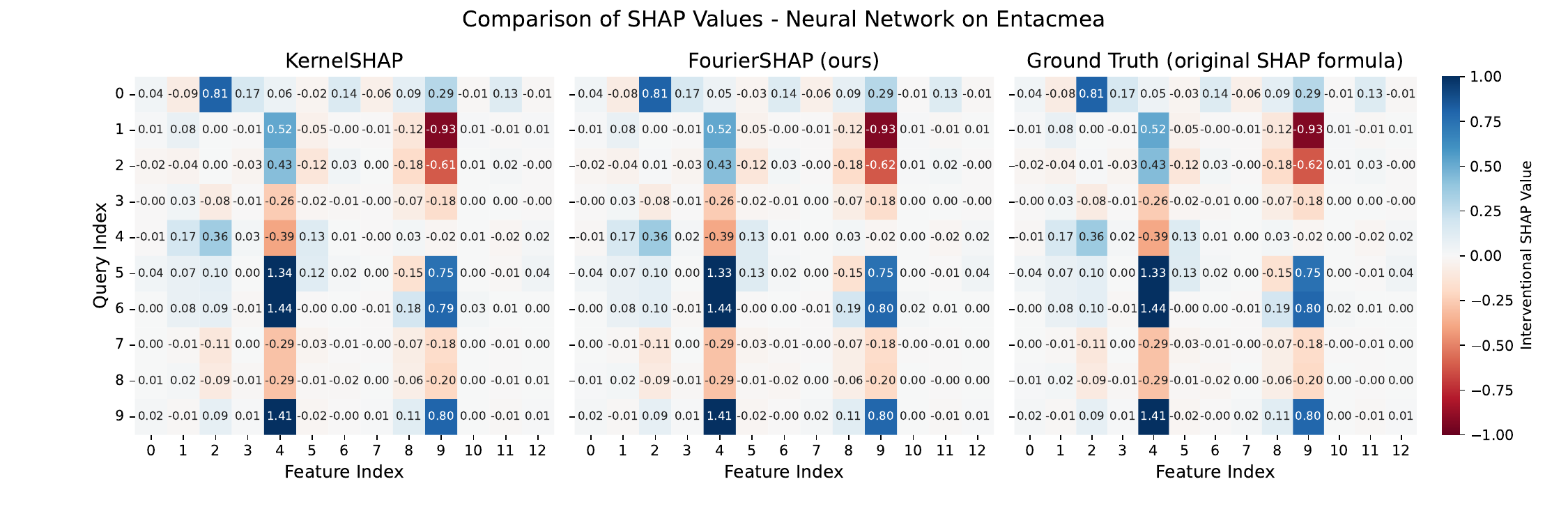}
    \caption{From left to right, 1- SHAP values generated by \textsc{KernelSHAP}, 2- SHAP values generated by FourierSHAP (ours), 3- and the ground truth SHAP values computed by the original exponential SHAP formula~Equation\ref{eqn:shap} on the Entacmaea dataset. In 3, Computation of ground truth SHAP values using the exponential formula is possible due to the dataset containing all $2^{13}$ possible boolean feature vectors. 10 query points and a background dataset points are chosen at random and are of size 10. This figure shows that both KernelSHAP and our method compute exact ground truth SHAP values.}
\label{fig:ground_truth}
\end{figure*}

In order to check if \textsc{KernelShap} values have converged to ground truth interventional SHAP values, 
we update the number of subset samples \textsc{KernelShap} uses to generate SHAP values for each instance. 
Looking into \textsc{KernelShap} repo \footnote{\url{https://github.com/shap/shap/blob/master/shap/explainers/\_kernel.py}}, this is the default number of samples in the code:
\begin{verbatim}
self.nsamples = 2 * self.M + 2**11 
\end{verbatim}
Here, \verb|self.M| is the number of indices where the instance is different from the background dataset.

To understand the convergence dynamics of \textsc{KernelShap}, we multiply this number by multiple ``sample factor''s and run the algorithm.
This allows us to experiment with the number of subsets sampled, and find a sample factor that ensures convergence while avoiding unnecessary extra computation. We test sample factors in the set $\{0.02, 0.03, 0.04, 0.1, 1.0, 2.0, 10.0\}$. 

For instance, for GB1 dataset, we compute the $R^2$ score of the \textsc{KernelShap} SHAP values with different sample factors to sample factor 10 (as the ground truth).
\begin{table}[h]
    \centering
    \begin{tabular}{|c|c|c|}
        \hline
        \textbf{Sample Factor} & \textbf{Runtime} & \textbf{$R^2$ w.r.t Ground Truth} \\
        \hline
        0.02 & 4.031 & 0.902 \\
        0.03 & 4.274 & 0.934 \\
        0.04 & 4.530 & 0.967 \\
        0.1  & 5.399 & 0.991 \\
        1.0  & 19.452 & 0.999 \\
        2.0  & 34.451 & 1.0 \\
        10.0 & 160.251 & 1.0  ($R^2$ w.r.t itself trivially=1)\\
        \hline
    \end{tabular}
    \caption{Table of sample factor, runtime, and $R^2$ with respect to sample factor 10, on GB1 dataset.}
    \label{tab:sample_factor_runtime}
\end{table}
For this dataset, the default setting, i.e., sample factor $=1$, seems good enough to make sure we are producing ground truth values and that the algorithm has converged. This is the general procedure we use for all datasets.  
\section{Limitations \& Future Work}
\label{app:limitations}
\subsection{Extending to continuous features}
A limitation of \textsc{FourierShap} is that it currently does not natively support continuous features and they have to be handled by quantization into categorical features. However, there is value in computing efficient SHAP values for models with continuous features and this is an important potential future work. In the following, we discuss how we think such an extension would work for the two classes of models we extensively covered in this work, namely trees and MLPs.

\textbf{Trees in the white-box setting.} Trees can be seen as inherently discrete structures even though they perfectly work for continuous features. By setting a threshold, i.e., a continuous number to define a split of the node, all trees are inherently in fact binary. This way of thinking gives one very simple but crude extension of the current framework: To assign to each node of the tree a binary feature specifying whether a certain continuous feature is bigger or smaller than its threshold. This would increase the feature dimension of the problem to the number of nodes in the tree in the worst case scenario. With the careful design of the transformation of continuous features into node-based binary features, this could be a potential extension for which the current work can lay a foundation. 

A second and more principled and less crude way to think about the case of trees is to not use a Fourier transform that we are using now which is over $Z_2 \times Z_2 \times ... Z_2$ (n times where n is the number of features). But rather use a transform over $Z_{K_1}, \times, Z_{K_n}$ where $K_i$ is the number of distinct thresholds feature $i$ has in the tree. Coming up with a closed form solution for the SHAP values in this Fourier basis is an interesting question that we also plan to look at. It is not too far-fetched to think that we can also find a closed form solution to this discrete Fourier basis. This would perfectly handle the case of any tree. Note that by a recursive formula one can readily compute the Fourier transform of trees as well. 

\textbf{MLPs.} Regarding the case of MLPs with continuous input features, one can not simply resort to the discrete Fourier transform of type $Z_k \times \dots Z_k$ to perfectly capture this structure as it is no longer inherently discrete (like trees). This is a slightly tricky scenario because it requires us to characterize the relationship of continuous and discrete Fourier transforms of neural networks functions. In classical signal processing there are results which characterize what level of granularity in discretization is required to reconstruct continuous transform with the discrete one for a given level of error. These results depend on the Spectrum of the continuous transform. For example the case of super imposition of sinusoids, where the spectrum has a bounded domain give rise to the Nyquist rate etc. 

For Neural network functions, if one had clear theoretical results on the spectral behavior of continuous transforms of neural networks it is not too hard to come up with conditions on the level of quantization required for the discrete transform to approximate them. One result that computes the spectrum appears over a multi dimensional inputs is (Theorem 1 of \cite{rahaman2019spectral}) but we are not aware of any spectral bias bounds on the continuous spectrum derived from this equation. However, for Neural networks with a single input dimension, bounds have been derived on the spectrum~\cite{cao2019towards} (in addition to the empirical results of \cite{rahaman2019spectral}). This intuitively means that single input Neural networks provably have a tendency to approximate functions with lower frequency sinusoids in the continuous domain. This implies that a discrete Fourier transform over $Z_K$ can approximate these function correctly for a sufficiently large discretization parameter K. This gives hope that a empirically a discrete Fourier transform of type $Z_K \times \dots Z_K$ for a sufficiently large $K$ would also approximate neural network function with higher dimensional inputs for a large enough discretization parameter K. Therefore, an extension of our closed-form solution for SHAP to $Z_K \times \dots Z_K$ transform could be a potential direction for the future work.

\subsection{Conditional SHAP values}
Another direction for future work is extending this work to compute conditional SHAP values. In this work, we only cover interventional SHAP values (the difference of interventional and conditional SHAP versions is discussed in detail in Appendix~\ref{subsec:shapley_literature_review}). 

In model-based methods like \textsc{TreeSHAP}~\cite{lundberg2020local}, one can use the tree itself to approximate conditional distributions of the data in a tractable way. One idea for extending this work is to attempt to tractably find a representation of the conditional distribution of the data using a sparse Fourier approximation of the predictor. Next, computing conditional SHAP Values using this conditional approximation.

\end{document}